\documentclass[journal]{IEEEtran}
\usepackage{microtype}
\usepackage{comment}
\usepackage{graphicx}
\usepackage{subfigure}
\usepackage{booktabs} % for professional tables
\usepackage{amsthm}
\usepackage{amsfonts,amssymb}
\usepackage{multirow}
\usepackage{enumitem}
\usepackage{caption}
\usepackage{algorithm}
\usepackage{algpseudocode}
\usepackage{amsmath}
\theoremstyle{definition}

\theoremstyle{plain}

\theoremstyle{plain}
\newtheorem{theorem}{Theorem}
\theoremstyle{remark}

\theoremstyle{plain}
\newtheorem{lemma}{Lemma}
\theoremstyle{plain}

\theoremstyle{plain}
\usepackage{xcolor}

\usepackage{amsmath}

\usepackage{hyperref}
\ifCLASSINFOpdf
\else
\fi
\begin{document}

\title{Agglomerative Neural Networks for Multi-view Clustering}


\author{Zhe~Liu,
        Yun~Li,
        Lina~Yao,
        Xianzhi~Wang,
        and~Feiping~Nie% <-this % stops a space
%\thanks{Z. Liu, Y. Li, and L. Yao are with the School of Computer Science and Engineering, University of New South Wales, Sydney, NSW 2052, Australia (e-mail: zhe.liu1@student.unsw.edu.au; yun.li5@student.unsw.edu.au; lina.yao@unsw.edu.au).}
% <-this % stops a space
%\thanks{X. Wang is with the School of Computer Science, University of Technology Sydney, Sydney, NSW 2007, Australia (e-mail: xianzhi.wang@uts.edu.au).}
% <-this % stops a space
%\thanks{F. Nie is with the School of Computer Science, Northwestern Polytechnical University, Xi’an 710072, China, and also with the Center for OPTical IMagery Analysis and Learning (OPTIMAL), Northwestern Polytechnical University, Xi’an 710072, China (e-mail: feipingnie@gmail.com).}
}




\maketitle

\begin{abstract}
Conventional multi-view clustering methods seek for a view consensus through minimizing the pairwise discrepancy between the consensus and subviews. However, the pairwise comparison cannot portray the inter-view relationship precisely if some of the subviews can be further agglomerated. To address the above challenge, we propose the agglomerative analysis to approximate the optimal consensus view, thereby describing the subview relationship within a view structure. We present Agglomerative Neural Network (ANN) based on Constrained Laplacian Rank to cluster multi-view data directly while avoiding a dedicated postprocessing step (e.g., using $K$-means). We further extend ANN with learnable data space to handle data of complex scenarios. Our evaluations against several state-of-the-art multi-view clustering approaches on four popular datasets show the promising view-consensus analysis ability of ANN. We further demonstrate ANN's capability in analyzing complex view structures and extensibility in our case study and explain its robustness and effectiveness of data-driven modifications.
\end{abstract}


% \fi
\IEEEpeerreviewmaketitle

\section{Introduction}
Clustering is a type of unsupervised machine learning techniques that partition data points into groups based on feature similarity.
Conventional clustering algorithms \cite{nie2016constrained,vidal2011subspace,kriegel2011density,kleindessner2019guarantees,oyelade2010application,peng2020deep,ergul2016clustering} are mostly single-view algorithms, which only consider single-source datasets.
Therefore, they cannot leverage complex view structures and cannot competently handle complex scenarios.
However, many real-world objects contain complex view structures, where each subview carries some unique information and the relationships existing between views may provide complementary information. 
For instance, when analysing a speech, the fusion of text data, voice data, and the relationships between them is more informative than a single view.
Thus, it calls for a multi-view clustering method that can leverage view structures effectively.
%Since most , we could collect the leveraged information from diverse views.
%Generally, each subview can carry some unique information,
%independent from other views, 
%and there exist relationships between views. 
%, e.g., image information is composed of LBP and HOG. 
%It is theoretical to combine HOG + LBP as the image domain, and then aggregate domain information from text + image + voice to achieve the overall information.
%Since single-view algorithms could not utilize the complex view structure information, multi-view clustering methods that can utilize the view structure are highly desirable.

Currently, multi-view clustering \cite{aaai2017,abnomral2005,ijcaisw,peng2019comic} usually comprises two steps to utilize and fuse view information: geometric consistency (GC) learning and cluster assignment consensus (CAC) learning. GC aims to capture the intrinsic similarity information within a single view; CAC aims to approximate the consensus view, which can combine the diverse similarity information from the subviews in a unified view. Although the existing research has achieved remarkable progress in computer vision, neural language processing and many other fields, there still exist challenges in GC learning and CAC learning.

The first challenge is that most current research fails to combine the advantages of two main kinds of GC learning: compactness-based methods and connectivity-based methods. Compactness-based methods look for distinct representations based on the similarity information (e.g., the eigenvectors of affinity matrix) to embed points \cite{coreg2011,cotraining2011,cikm17_mcge,zhao2018incomplete}. Although they are good at extracting informative embeddings, they need postprocessing (e.g., $K$-means) to obtain the clustering results, which may impair the consistency between the learned representations and the final clustering results.
%To obtain the clustering results, these compactness-based GC analyses need further processing (e.g., $K$-means) to partition data points into groups.
%However, the postprocessing operation may impair the consistency between the learned representations and the final clustering results.
%To solve this problem, 
In comparison, connectivity-based GC methods project data and encode the similarity information in connection graphs directly.
%Such methods are good in keeping the clustering consistency \cite{aaai2017,ijcaisw,mvgl,peng2019comic}, as 
The connection graphs can assign cluster labels according to the connected components to ensure the consistency between latent representations and the clustering \cite{aaai2017,ijcaisw,mvgl,peng2019comic}. However, such methods may lose information while embedding raw data to connection graphs directly. It is necessary to propose an extensible data-specific framework that embraces the latent representation learning and clustering consistency simultaneously.

\begin{figure}
    \centering
    \includegraphics[width=\linewidth]{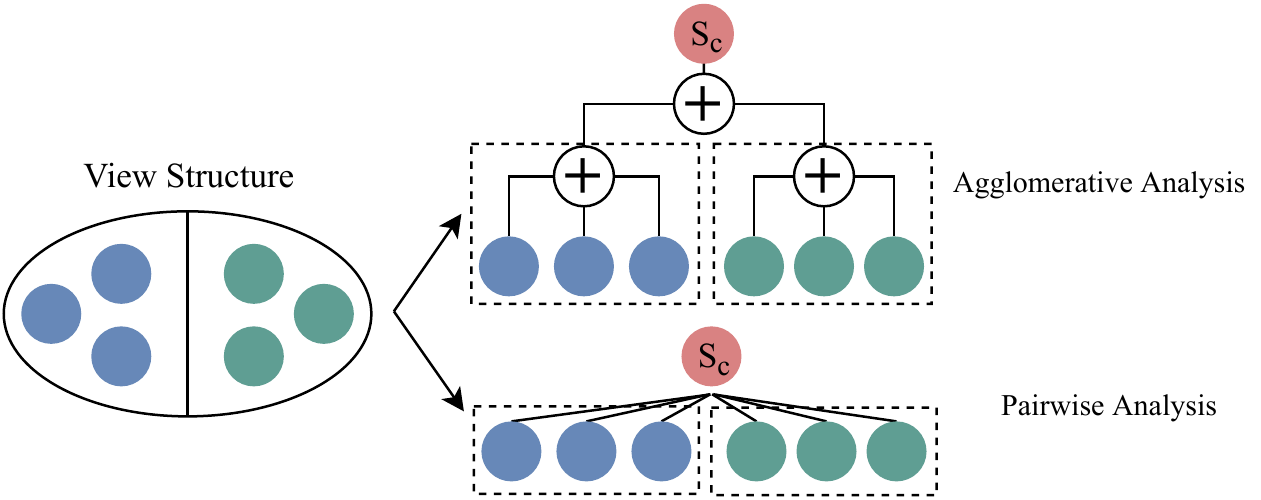}
    \caption{Comparison of pairwise analysis and agglomerative analysis. Given a view structure that contains six independent subviews from two domains (in green and blue, respectively), where each circle represents the information of a subview, pairwise analysis approximates a consensus view $S^{(c)}$ through comparing $S^{(c)}$ with each subview without leveraging the structural information, while agglomerative analysis obtains $S^{(c)}$ by agglomerating subviews following the view structure.}
    \label{ANN_intro}
\end{figure}

The second challenge is that current CAC research fails to explore view relationships. Most CAC analysis research is inspired by Kumar et al. \cite{coreg2011,cotraining2011}.
%, and the multi-view structure analysis mainly relys on pairwise subview comparison~\cite{peng2019comic,aaai2017,ijcaisw,mvkdr,scikit-learn,bmvc}.
%To be specific, these CAC studies aim to find consensus cross views.
The corresponding algorithms rely on pairwise subview comparison~\cite{peng2019comic,aaai2017,ijcaisw,mvkdr,scikit-learn,bmvc} to minimize the discrepancy between the consensus view and each subview.
%Considering that the pairwise comparison cannot portray complex view structures that consist of subdivisions,
%To solve this problem,
%we propose agglomerative analysis and illustrate its difference from pairwise analysis in Fig.~\ref{ANN_intro}.
Since the pairwise subview comparison methods only analyze the subview independently,
%in multi-view analysis and omit 
they cannot utilize the structural relationship
%(left part of Fig.~\ref{ANN_intro}) 
when handling complex hierarchical view structures. 
%the conventional view analysis
%cannot precisely portray the view structure and lack the capability of utilizing the structured view information when handling complex hierarchical view structures. 

To address the challenges above, we propose agglomerative consensus analysis to portray complex view structures and utilize view relationships. Its difference from the pairwise analysis is illustrated in Fig.~\ref{ANN_intro}.
Based on the agglomerative consensus analysis, we further propose Agglomerative Neural Network (ANN) to embrace latent representation learning and connectivity-based analysis in a multi-view format.
Considering the extensibility of neural networks, we further implement agglomerative consensus analysis through an Agglomerative Neural Network with Learnable Data space (ANNLD). ANNLD introduces data-specific learnable projection to improve the data distribution in ANN.

In summary, we make the following contributions:
\begin{itemize}
    \item We first propose an agglomerative consensus analysis as a unified framework of latent representation learning and Laplacian rank constrained multi-view clustering, which takes advantages of both compactness-based and connectivity-based GC analysis. 
    %learn a projection automatically to improve the raw data distribution.
    %, which indicates the excellent extensibility of our network on diverse scenarios.
    \item We 
    %further specify the framework components and realize 
    present an Agglomerative Neural Network (ANN) to implement agglomerative consensus analysis and adopt its data-specific extension, Agglomerative Neural Network with Learnable Data space (ANNLD), to learn more discriminative projection from entangled data across different views and subviews. 
    \item Our experiments on four commonly used multi-view datasets show ANN's superiority and robustness in analyzing different multi-view datasets. Our experimental results on Survey dataset demonstrate the excellent performance of ANNLD and extensibility of agglomerative consensus analysis.
\end{itemize}

\section{Related Work}
%The existing multi-view clustering algorithms can be sorted into connectivity-based and compactness-based methods. 
The connectivity-based clustering algorithms focus on finding a connection graph to represent raw information and thus directly obtain the clustering results. Therefore, connectivity-based methods may better preserve clustering consistency. Nie et al. \cite{swmc,aaai2017} proposed to learn the consensus connection graph by comparing it with each subview's affinity matrix or distance matrix, which used self-weighted subviews to reduce the process of optimization. Since Nie researched on fixed affinity matrices or distance matrices of subviews, Zhan et al. \cite{mvgl} further proposed learnable subview connection graphs and then fused the subviews to obtain the consensus view. Wang et al. \cite{wang2019gmc} proposed to find a fusion view to represent the multi-view information; they further approximate the fusion view by combining weighted subviews under constrained Laplacian rank. Huang et al. \cite{peng2019comic} learned connection graphs from fixed projected data space and aimed to eliminate the mismatching problem across different views. 

The compactness-based methods encoded the clustering information in the eigenvector matrix (or latent representations) and used the postprocessing methods (e.g., $K$-means) to cluster the learned representations. Kumar et al. \cite{coreg2011,cotraining2011} optimized the eigenvector matrix to contain the clustering indicators and searched for the consensus subview or cross-view by pairwise comparison. Zhang et al. \cite{bmvc} applied matrix factorization and constrained the latent codes to learn the consensus representations in a binary structure. Zhang et al. \cite{zhang2015low} proposed to learn complementary information from multiple views via constrained tensors. The tensors captured the high order correlation underlying views to reduce cross-view redundancy of the learned subspace representations. Zhou et al. \cite{zhou2019multiple} introduced neighbor-kernel-based algorithm that utilized the intrinsic neighborhood structure to preserve the block diagonal structure and to strengthen the robustness against noise and outliers. The algorithm fused these base neighbor kernels to extract a consensus representation through subspace learning.

Moreover, some deep-learning-based algorithms \cite{andrew2013deep,wang2015deep,wei2019multi} extended the previous work with a non-linear relationship and learned the canonical correlation between views. Wu et al. \cite{wu2019essential} introduced Markov-chain-based spectral clustering method to find the essential tensor of high order correlation representation. Zhang et al. \cite{zhang2018generalized} proposed general relationship learning based on neural networks to learn the pairwise relationship between views and thus obtain the fused view. Huang et al. \cite{huang2019multi} utilized Siamese network and applied orthogonal constraint to enable network performing local invariance learning and matrix decomposition, which further enhance the pairwise comparison based subspace learning. However, the above algorithms mainly relied on the pairwise comparison between subview and consensus view, so they failed to utilize the view structure information which may provide complementary cross-view information. Besides, few of the mentioned work combined connectivity-based learning and compactness-based in a unified neural network. 

Compared to the approaches above, our contributions in this work are two-fold. First, most of the existing approaches only utilize the simple view structure without subdivisions. In contrast, our method specifies more details in the view structure and is capable of dealing with complex hierarchical view structures with subdivisions of subviews. The proposed agglomerative consensus analysis can capture and portray the subdivision relationship when agglomerating subviews. Such agglomerative analysis is more theoretical and effective than conventional consensus analysis in utilizing view information. Second, most current research exclusively studies latent representation for view information, constrained Laplacian matrix, and cross/pairwise consensus view analysis. Little work has been done to incorporate them in a unified framework, and the limited existing studies fail to make a data-specific extension. In comparison, we propose a unified and extensible deep learning-based algorithm that can overcome the above deficiencies.

\section{Agglomerative Consensus Analysis}
\begin{figure*}
    \centering
    \includegraphics[width=\textwidth]{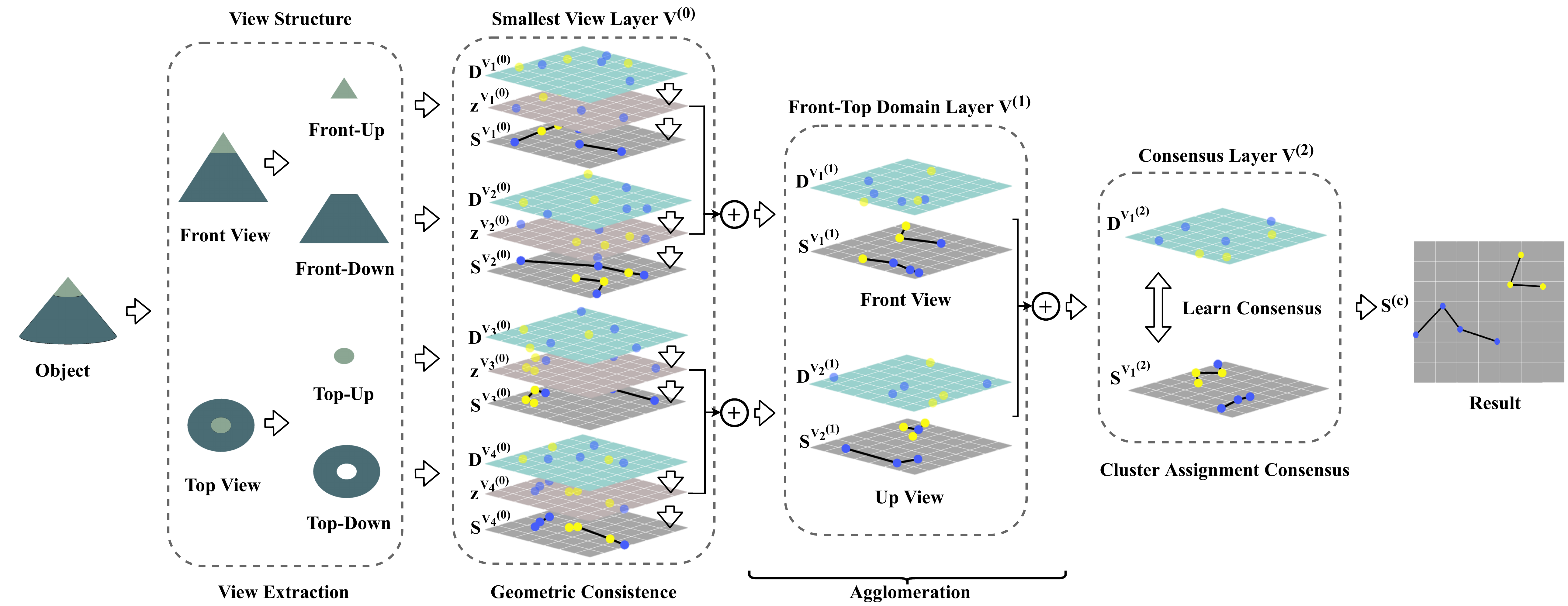}
    \caption{An illustration of the proposed
    %\textcolor{red}{ANN in the view structure composed of four subviews from two domains. Graphs with blue axis and grey axis denote distance matrices and the learned connection graphs, respectively. ANN first learns and converts subview information to connection graphs via geometric consistency. Then, it separately agglomerates distance matrices and connection graphs based on the view structure. Last, ANN moderates the discrepancy during the agglomeration and obtains the consensus-view connection graph as results.}\textcolor{blue}{
    ANN. We take an object to show a two-layer view structure consisting of four subviews from two domains. ANN learns the subview information $z^{(v)}$ from $D^{(v)}$ via geometric consistency, converts it into a connection graph $S^{(v)}$, and finally, agglomerates subviews by layers and minimizes the discrepancy in the consensus layer to obtain the clustering result $S^{(c)}$.}%}
    \label{ANN_flow}
\end{figure*}

%This section comprises two main parts: view structure declaration and agglomerative consensus analysis explanation. Since little work has explained the meaning of view structure, we first introduce and declare the view structure in multi-view datasets. According to the view structure, we further explain how the agglomerative consensus analysis agglomerate and obtain the consensus information in a general form.
This section introduces the methodology and theory to carry out the agglomerative consensus analysis framework for multi-view datasets. The full details of component realization in the neural network will be discussed in Section \ref{realization}.

The proposed agglomerative loss function comprises three terms:
\begin{equation}
    \mathcal{L}=\lambda\mathcal{L}_{sc}+\mathcal{L}_{gc}+\mathcal{L}_{cac}
\end{equation}
where $\mathcal{L}_{sc}$ constrains Laplacian matrix rank and controls the clustering convergence; $\lambda$ denotes a weighted parameter; $\mathcal{L}_{gc}$ and $\mathcal{L}_{cac}$ enable the model to learn the multi-view information. To be specific, $\mathcal{L}_{gc}$ encodes the subview information in latent representations by learning geometric consistency; $\mathcal{L}_{cac}$ keeps cluster assignment consistent across views.

\subsection{Constrained Laplacian Rank for Spectral Clustering}
%First, 
Constrained Laplacian Rank (CLR) loss $\lambda\mathcal{L}_{sc}$, which derives from the spectral clustering, has been a widely used tool to carry out clustering on datasets without any postprocessing.
Let $G=(X,S)$ be an undirected graph on $X=[n]$ and the connection graph $S$ be the corresponding edge set. We assume: for any two arbitrary sample point $X_{i}$ and $X_{j}$, $S_{ij}$ carries a non-negative weighted edge to encode the similarity intensity between the points. If there exists a edge between $X_{i}$ and $X_{j}$, $S_{ij}>0$; otherwise, $S_{ij}=0$. Specially, $\forall i\in [n], S_{ii}=0$. We keep the main diagonal element of connection graph equals 0 to ensure the graph is undirected. Further, let $Dg$ be the degree matrix, which is a diagonal matrix $Dg_{ii}=\sum_{j\in[n]}S_{ij},i\in[n]$.

Given $k\in \mathbb{N}$, spectral clustering aims to cut edges with minimum weights and partition $X$ into k clusters. Let $L_{S}=Dg-(S^{T}+S)/2$ denote the unnormalized Laplacian matrix, and spectral clustering solves the cutting problem by minimizing following loss function~\cite{von2007tutorial}:
\begin{equation}
    \begin{gathered}
    \min_{H\in \mathbb{R}^{n\times k}}Tr(H^{T}L_{S}H)\\
    s.t.\quad H^{T}H=I_{k}.
    \end{gathered}
\end{equation}
where $I$ is the identity matrix and $Tr$ is the trace of matrix. A common solution to $H$ is the eigenvector matrix $F$, which consists of the orthonormal eigenvectors corresponding to the $k$ smallest eigenvalues of $L_{S}$~\cite{lutkepohl1996handbook}. 

Further, we utilize the relationship between eigenvalues and the graph connected components, which can constrain the Laplacian matrix, to directly get clustering results.
\begin{theorem}\label{ComponentNumber}
The multiplicity of the eigenvalue 0 of the Laplacian matrix $L_{S}$ (non-negative) equals the number of the connected components in the connection graph $S$.
\end{theorem}
Theorem \ref{ComponentNumber} indicates that 0 eigenvalues' multiplicity equals the cluster number. If there exist $k$ clusters on $X=[n]$, the rank of corresponding Laplacian matrix $L_{S}$ should be $n-k$. Let $\sigma_{i}(L_{S})$ be the $i$th smallest eigenvalues of $L$. According to Ky Fan's Theorem (Fan 1949), we can relate $F$ to the Laplacian matrix rank
\begin{equation}\label{CLRLOSS}
    \begin{gathered}
    \mathcal{L}_{sc}=\sum_{i=1}^{k}\sigma_{i}(L_{S})=\min_{F,S}Tr(F^{T}L_{S}F)\\
    s.t. \quad F^{T}F=I, rank(L_{S})=n-k
    \end{gathered}
\end{equation}
where $L_{S}$ denotes the corresponding Laplacian matrix of connection graph $S$. The loss function aims to regularize the Laplacian matrix rank to be exactly $n-k$ by minimizing the sum of smallest $k$ eigenvalues to 0. Then, the connection graph will establish $k$ clusters and thus the clustering labels can be directly obtained from $S$.

\subsection{Agglomerative Multi-view Analysis}
This section first clarifies the view structure of the multi-view dataset and then introduces the proposed agglomerative multi-view analysis. The multi-view analysis comprises two parts: $\mathcal{L}_{gc}$ encodes distance information in latent representations by learning GC; $\mathcal{L}_{cac}$ minimizes the discrepancy between the projected connection graph and agglomerated raw information to achieve CAC. 

We define $\sim$ as the subview relationship and $\in$ as the belonging relationship. Let $V=\{V^{(0)},V^{(1)},...,V^{(m)}\}$ be a m-layer view structure and $V^{(i)}_{j}$ be the $j$th view in $i$th layer, where $V^{(i)}$ represents the $i$th layer view set. Specially, $V^{(0)}$ consists of the smallest independent subviews and $V^{(m)}$ should only have one member $V^{(m)}_{1}$ to represent the consensus view i.e. the final combined view for the multi-view dataset. Let $v$ be an arbitrary view, say $V^{(i)}_{j}$. For the $0$th layer, we let $\forall v\in V^{(0)}$ be an independent subview belonging to the 0th layer;
%e.g., HOG and LBP features; 
for any subsequent layer $\forall v\in V^{(i)}$ ($i \geq 1$), we denote the agglomerated view by its corresponding subviews, i.e., $v=\{V^{(i-1)}_{j}:V^{(i-1)}_{j} \sim v\}$, where $\sim$ denotes that $V^{(i-1)}_{j}$ is one of the related subview of $v$ and the relationship is predefined in the dataset. 
%the agglomerated view by its corresponding subviews, 
%As an example, an image-view is agglomerated from HOG and LBP features.
%, where $v\in V^{(i)}(i \geq 1)$ is composed of the corresponding former layer's views. 
For the ease of illustration, we signify $v^{'}_{j}$ as the $j$th subview of $v$ and $v=\{v^{'}_{j}:v^{'}_{j} \sim v\}$.

%\textcolor{red}{To help understand the definition of the view structure, we start with a ordinary multi-view dataset comprises $n$ subviews without any subdivisions. The view structure can be written as $V=\{V^{(0)},V^{(1)}\}$, where $V^{(0)}=\{V^{(0)}_{1},V^{(0)}_{2},...,V^{(0)}_{n}\},V^{(1)}=\{V^{(1)}_{1}\}$. The $V^{(1)}_{1}$ is the fusion of all the subviews in $V^{(0)}$, so $V^{(1)}_{1}=\{V^{(0)}_{1},V^{(0)}_{2},...,V^{(0)}_{n}\}$. }

%\textcolor{red}{Furthermore, we consider the condition with subdivisions. Recalling the view structure in Fig.~\ref{ANN_intro}, a two-layer view structure $V=\{V^{(0)},V^{(1)},V^{(2)}\}$ consists of two view domains and each domain can be further subdivided into three independent subviews. We number the smallest independent subviews in the figure from the left to the right and denote $V^{(0)}=\{V_{1},V_{2},...,V_{6}\}$. We can summarize the domain views by $V^{(1)}=\{V^{(1)}_{1},V^{(1)}_{2}\}, V^{(1)}_{1}=\{V^{(0)}_{1},V^{(0)}_{2},V^{(0)}_{3}\},V^{(1)}_{2}=\{V^{(0)}_{4},V^{(0)}_{5},V^{(0)}_{6}\}$ and annotate the consensus view by $V^{(2)}_{1}=\{V^{(1)}_{1},V^{(1)}_{2}\}$. The superscript helps differentiate the view-level of the combined view (i.e., $V^{(0)}_{i}, V^{(1)}_{i}, V^{(2)}_{i}$ represent a view in the basic-level, domain-level and consensus-level, respectively), and the subscript assists distinguishing the views in the same level (i.e., $V^{(1)}_{1},V^{(1)}_{2}$ denote two independent domain views).}

Given a {\em m}-layer view structure $V$ as above, we assume latent representation $z^{(v)}$ encodes the distance information for each subview $v\in V^{(0)}$. Suppose the corresponding raw information of view $v$ is $D^{(v)}$ (i.e., distance matrix), latent representations $z^{(v)}\in \mathbb{R}^{n\times n}$ minimizes the below loss function to learn GC:
\begin{equation}\label{gc loss}
    \begin{gathered}
    \mathcal{L}_{gc}=\min_{Z}\sum_{v\in V^{(0)}}\left (\sum D^{(v)}\circ z^{(v)}+\left \| z^{(v)} \right \|_{F}^{2}\right )\\
    s.t. \quad Z:=\{z^{(v)}:v\in V^{(0)}\}
    \end{gathered}
\end{equation}
where $\circ$ means Hadamard product. $Z$ denotes the target distance representation set and $z^{(v)}=((z^{(v)})^{T}+z^{(v)})/2$ ensures the distance representations being symmetric.

The first term enables latent representations to encode the raw distance information. The second term is a penalty term to prevent $z_{ij}\rightarrow -\infty$.

We further explain the agglomerative consensus analysis theory. Our goal is to acquire a consensus view $S^{(c)}$ concatenating the multiple view information. It is intuitive to utilize the hierarchical view structure to approximate the fused view layer by layer. Since the fused view information should be related to all the corresponding subviews, we propose our agglomerative analysis by assuming there exists a function $\gamma^{(v)}$ which projects the corresponding subviews to the fused views $v\in V^{(i)}(i \geq 1)$. Given the learned latent representation set $Z:=\{z^{(v)}:v\in V^{(0)}\}$, we assume an activation function $\mathcal{C}(z)\rightarrow S$ can convert representation to an normalized connection graph. Then, we define:
\begin{equation}\label{agglomaration}
S_{v}=\begin{cases}
\mathcal{C}(z^{(v)}) & \text{ if } v\in V^{(0)} \\ 
\gamma^{(v)}(\{S^{(v^{'}_{j})}:v^{'}_{j}\sim v\}) & \text{ if } v\in V^{(i)},i \geq 1
\end{cases}
\end{equation}
where $v^{'}_{j}$ is the $j$th corresponding subview $V^{i-1}_{j}$ to compose $v$;
$\gamma$ denotes the agglomeration operation. The connection graphs in $V^{(0)}$ are calculated from latent representations and the other graphs are achieved by agglomeration. 

We propose an activation function $\mathcal{C}$ to convert latent representations and regularize agglomerated graphs. The activation function ensures that $S$ is a normalized connection graph.
Section \ref{architecture details} elaborates on the activation function and the agglomeration.

Given the consensus connection graph $S^{(c)}$ for a m-layer view structure $V$, we sort the general agglomerated form based on Eq. (\ref{agglomaration}):
\begin{equation}\label{sc formula}
\begin{gathered}
\{S^{(v)}:v\sim V^{(i)}\}=\{\gamma^{(v)}(\{S^{(v^{'}_{j})}:v^{'}_{j}\sim v\}):v \sim V^{(i)}\} \\
S^{(c)}=\gamma^{V^{(m)}_{1}}(\{S^{(V^{(m-1)}_{j})}:V^{(m-1)}_{j}\sim V^{(m)}_{1}\})
\end{gathered} 
\end{equation}

Since the last layer will combine all the views of the former layer as a unified view, we let $V^{(m-1)}_{j}$ denote the $j$th subview of the consensus view; $1\leq i < m$.

We hope the consensus connection graph $S^{(c)}$ can learn subview information evenly. Therefore, we also agglomerate distance matrix to achieve the agglomerated consensus raw information $D^{(c)}$:
\begin{equation}\label{dv formula}
\begin{gathered}
\{D^{(v)}:v\sim V^{(i)}\}=\{\sum_{v^{'}_{j}\sim v}w_{d}^{(v^{'}_{j})}D^{(v^{'}_{j})}:v \sim V^{(i)}\}\\
D^{(c)}=\sum_{V^{(m-1)}_{j}\sim V^{(m)}_{1}}w_{d}^{(V^{(m-1)}_{j})}D^{(V^{(m-1)}_{j})}
\end{gathered}
\end{equation}

where 
%\textcolor{red}{$w_{d}^{(j)}=1/|v|$, i.e., the related subview quantity.}
$w_{d}^{(v^{'}_{j})}=1/|v|,w_{d}^{(V^{(m-1)}_{j})}=1/|V^{(m)}_{1}|$, i.e., the corresponding subview quantity of an agglomerated view.

The consensus raw information $D^{(c)}$ evenly fuses the subview information based on the view structure, and the clustering assignment consensus problem will be converted to minimize the discrepancy between $S^{(c)}$ and $D^{(c)}$. Similarly, we can figure out the loss function for CAC:
\begin{equation}\label{ann cac}
\begin{gathered}
\mathcal{L}_{cac}=\min_{Z,\tau}\sum D^{(c)}\circ S^{(c)}+\left \| S^{(c)} \right \|_{F}^{2}\\
s.t.\quad Z:=\{z^{(v)}:v\in V^{(0)}\},\tau:=\{\gamma^{(v)}:i \geq 1, v\in V^{(i)}\}\\
S^{(c)}=\underbrace {\{{\gamma^{v\in V^{(m)}}(\cdots\{\gamma^{v\in V^{(1)}}(}}_{m \ layers} \{\mathcal{C}(z^{v^{'}_{j}}):v^{'}_{j}\sim v\})\})\}
\end{gathered}
\end{equation}
%\textcolor{red}{where $\mathbf{1}$ denotes a full 1 column vector.}
where $S^{(c)}$ is agglomerated from the converted latent representations layer by layer.

Since $S^{(c)}$ is driven by the agglomerative operation and latent representations, the loss function optimizes $Z,\tau$ to achieve an optimal $S^{(c)}$.
Specifically, $\mathcal{L}_{cac}$ approximates such a projection to agglomerate the subviews that can balance making the cross-view solution distribute evenly and optimizing the solution for each subview.

\subsection{Convergence Analysis}
To cluster raw data into $k\in \mathbb{N}$ clusters, $\mathcal{L}$ converges when $rank(L_{S^{(c)}})=n-k$. Consider that $\lambda$ is large enough, $\mathcal{L}\approx \lambda Tr(F^{T}L_{S^{(c)}}F)$. Note that $\forall i, \sigma_{i}(L_{S^{(c)}})\geq 0$, the optimal solution $S^{(c)}$ will let the smallest $k$ eigenvalues be zero. 
\begin{lemma}\label{lemma1}
For every vector $f\in \mathbb{R}^{n}$, $f^{T}Lf=\frac{1}{2}\sum_{i,j=1}^{n}S_{ij}(f_{i}-f_{j})^{2}$.
\end{lemma}
\vspace{-4mm}
\begin{proof}
The Laplacian matrix definition ensures
\[
\begin{split}
f^{T}Lf&=f^{T}Dgf-f^{T}Sf=\sum_{1}^{n}Dg_{i}f_{i}^{2}-\sum_{i,j=1}^{n}f_{i}f_{j}S_{ij}\\
&=\frac{1}{2}(\sum_{i=1}^{n}Dg_{i}f_{i}^{2}-s\sum_{i,j=1}^{n}f_{i}f_{j}S_{ij}+\sum_{j=1}^{n}Dg_{j}f_{j}^{2})\\
&=\frac{1}{2}\sum_{i,j=1}^{n}S_{ij}(f_{i}-f_{j})^{2}
\end{split}  
\]
\end{proof}
\vspace{-4mm}
According to Lemma \ref{lemma1}, we can directly figure out
\begin{equation}
    Tr(F^{T}L_{S^{(c)}}F)=\sum_{1}^{k}\sigma(L_{S^{(c)}})=\frac{1}{2}\sum_{i,j=1}^{n}(F_{ij}-F_{ji})^{2}S^{(c)}_{ij}
\end{equation}

Therefore, $\mathcal{L}\approx \frac{\lambda}{2}\sum_{i,j=1}^{n}(F_{ij}-F_{ji})^{2}S^{(c)}_{ij}$. According to the chain rule in neural network, $\mathcal{L}$ keeps monotonically decreasing $S_{ij}$ unless $F_{ij}=F_{ji}$. Since only the 0 eigenvalue's corresponding eigenvector meets that $\forall i,j, f_{i}=f_{j}$, $\mathcal{L}$ optimizes $Z$ and $\tau$, which drive the optimization of $S^{(c)}$, and keeps cutting the edges to reduce the corresponding eigenvalues until the constrained Laplacian rank $rank(L_{S^{(c)}})=n-k$ can be established. Thus, $\mathcal{L}$ tends to converge.

\section{Agglomerative Neural Network}\label{realization}
This section introduces Agglomerative Neural Network (ANN) and its extended version, Agglomerative Neural Network with Learnable Data space (ANNLD), as well as their optimization methods.
%the detailed network operations and architectures to realize our proposed method. First, we show the network architecture to realize Agglomerative Neural Network (ANN) for a general situation. To further show the extensibility of our network, we propose Agglomerative Neural Network with Learnable Data space (ANNLD) under a complex scenario that original data points lack a significant difference in the raw data space. ANNLD poses a projection to transform the raw input data and optimizes the projection to assist clustering data points. Then, we reveal the matrix initialization and optimization for our networks.

\subsection{Agglomerative Neural Network}\label{architecture details}
The agglomerative consensus analysis relies on the agglomeration operation and constructs the consensus view layer by layer. Since neural networks have a chain structure, we only need to declare the deep learning agglomeration operation as layers in the network. 
%Starting with most significant component of our network, 
Since conventional neural network only optimizes latent representations according to loss gradient, %without other constraint, so 
it may not keep the learned connection graph as normalized.
%, e.g., each row of connection graph may contain negative elements.
In this regard, we design an activation function $\mathcal{C}$ that regularizes input data to be a normalized connection graph:
\begin{equation}\label{Activation Function}
\mathcal{C}(x_{i})=\begin{cases}
\frac{\mathcal{P}\cdot x_{i}-x_{min}}{\mathcal{P}( \sum_{j\in x^{+}}x_{j}-x_{min})} & \text{ if } x_{i}\geq 0\\
\quad \quad \quad 0 &  \text{ if } x_{i}<0 
\end{cases}
\end{equation}
where $x_{i}$ denotes $i$th element of a row vector $x$. $x_{min},x^{+}$ are the minimum and positive element of vector $x$. $\mathcal{P}$ is a hyper-parameter that prevents the edge from vanishing after rescaling data. The activation function $\mathcal{C}$ plays three roles in the network. First, it regularizes the input vector into a standard affinity vector in the connection graph. It rescales the input vector to $[0,1]$ and lets the sum of elements equal 1. Second, $\mathcal{C}$ also holds inequality relationships between non-negative elements and keeps non-positive edges inactivated to accelerate the optimization. Last, $\mathcal{C}$ can prevent some trivial solutions from cutting one point as a cluster. $\mathcal{C}$ tends to keep the last positive edge of points during agglomeration, which assigns the last edge weight as 1.

Based on the activation function in Eq. (\ref{Activation Function}), ANN can make sure that the learned connection graphs are normalized after converting distance representation and agglomerating subview matrices in Eq. (\ref{agglomaration}). Then, we discuss the agglomeration operation. Following that each subview contributes varying importance to the consensus view \cite{aaai2017}, the agglomerated representation, which encodes connection information, is achieved by weighted linear transformation. Given an arbitrary view $v\in V^{(i)}(i\geq 1)$ and the corresponding subview $\{S^{(v_{1})},S^{(v_{2})},...,S^{(v_{i})}\}$, the connection graph $S^{(v)}$ can be agglomerated by
\begin{equation}\label{sigma}
    S^{(v)}=\gamma^{(v)}(\{S^{(v_{1})},S^{(v_{2})},...,S^{(v_{i})}\})=\mathcal{C}(\sum_{i}w^{(i)}S^{(v_{i})})
\end{equation}
where $w^{(i)}$ denotes a learnable parameter to represent $v_{i}$'s weight in the agglomeration.

According to Eq. (\ref{Activation Function}) and Eq. (\ref{sigma}), we can realize ANN based on agglomerative consensus analysis. Denote $W$ as the set of $w^{(i)}$ in agglomeration operation, and we can sort the loss function for ANN:
\begin{equation}
    \min_{Z}[\lambda\mathcal{L}_{sc}+\mathcal{L}_{gc}+\mathcal{L}_{cac}];\min_{W}[\lambda\mathcal{L}_{sc}+\mathcal{L}_{cac}];\min_{F}[\lambda\mathcal{L}_{sc}]
\end{equation}

\subsection{ANN with Learnable Data Space}

We extend ANN with Learnable Data space (i.e., ANNLD) to address the challenges posed by data with complex view structures.
%There exist many overlapping points that limit the discriminative capability of the raw Survey data, as shown in .
%The detailed illustration will .
Fig.~\ref{embedding} shows an example of such data (i.e., Survey data in Section~\ref{expe}), where the raw data are overlapped and belong to different views or subviews.
%where the raw data overlap with limited discriminative capability---
Such data confuse the distance matrix and further prevent the algorithm from clustering data correctly; also, the minor discrepancy between data slows down the convergence of neural networks' gradient descending.
%Due to the extensibility of neural network, we design data-specific ANNLD by posing data-driven projections to improve distribution.

Let $X$ be the data with complex view structures and $X_{ij}$ be the $j$th criterion score of the $i$th interviewee.
For each criterion $j$, ANNLD applies $tanh$, a commonly used activation function, to obtain a better dimension distribution. ANNLD learns an extra parameter $h_{j}$ to modify the $j$th criterion distribution:
\begin{equation}\label{Learnable h}
   X^{'}_{j}=\{tanh(h_{j}\cdot (X_{ij}-\overline{X_{j}})):X_{ij}\in X_{j}\}
\end{equation}
where $\overline{X_{j}}$ is the mean score of dimension $j$ to ensure projected data ranging from -1 to 1; $h_{j}$ denotes a learnable parameter which controls the distribution shape of projected data space; $X^{'}_{j}$ denotes the projected feature dimension. 

Then, ANNLD will take the projected $X^{'}$ to replace the ordinary $X$ for further multi-view analysis. Due to the chain rule in the neural network, ANNLD will optimize $h$ to make the projected data space $X^{'}$ easier to be clustered. The optimized data space $X^{'}$ will be further discussed in Section \ref{DataSpace}.

Though we have used a projection to ease the overlapping problem, the small initial discrepancy may still make the optimization of projection and consensus analysis slow at the beginning. Thus, we add a bias parameter $b^{(v)}\in \mathbb{R}^{n\times n}$ in each view to assist dropping edges and to accelerate the optimization. 

Given an arbitrary view $v\in V^{(i)}(i\geq 1)$ and the corresponding subview $\{S^{(v_{1})},S^{(v_{2})},...,S^{(v_{i})}\}$, we could agglomerate the new connection graph $S^{(v)}$ by
\begin{equation}\label{Learnable b}
    S^{(v)}=ReLU(\mathcal{C}(\sum_{i}w^{(i)}S^{(v_{i})}+b^{(v)})
\end{equation}
where $ReLU$ denotes a Rectified Linear Unit that drops negative edges.

Then, we apply a penalty term of $b^{v}$ and obtain $\mathcal{L}_{cac}$ by
\begin{equation}\label{Learnable b 2} \mathcal{L}_{cac}=\sum D^{(c)}\circ S^{(c)}+\left \| S^{(c)} \right \|_{F}^{2}+\sum_{b^{(v)}\in B}\left \| b^{(v)} \right \|_{F}^{2}
\end{equation}

Since $\mathcal{L}$ will monotonically decrease $S^{(v)}$, the network will optimize $b^{(v)}$ to be negative, and ReLU will drop the edges and thus will accelerate the convergence of Laplacian matrix rank. 

Let $H$ and $B$ be the set of all learnable $h_{j}$ and $b^{(v)}$, we define the loss function for ANNLD based on Eq. (\ref{Learnable h}), Eq. (\ref{Learnable b}) and Eq. (\ref{Learnable b 2}):
\begin{equation}
    \min_{Z,H}[\lambda\mathcal{L}_{sc}+\mathcal{L}_{gc}+\mathcal{L}_{cac}];\min_{W,B}[\lambda\mathcal{L}_{sc}+\mathcal{L}_{cac}];\min_{F}[\lambda\mathcal{L}_{sc}]
\end{equation}

\subsection{Optimization}
$Z,H,W,B$ can be optimized by backpropagation of the gradient descent automatically, given the chain rule and good extensibility of neural networks. We update $F$ after each round of gradient optimization of the other variables. The optimal solution to $\min Tr(F^{T}L_{S}F)$ is a matrix composed of the $k$-smallest eigenvalues' corresponding eigenvectors, where $k$ denotes the target cluster number. Therefore, we update $F$ by the new eigenvectors of the smallest $k$ eigenvalues.

With the learnable parameters in the pruning edges, we also consider three different conditions to update $\lambda$: 1) when the current cluster number is smaller than the target number of $k$, we set $\lambda=min(\lambda_{max},2\cdot \lambda)$ to accelerate the cutting of edges; 2) when the current cluster number is greater than $k$, we set $\lambda=\lambda/2$ and restore the other parameters of the last turn to slow down edge-cutting speed; 3) when the current cluster number equals $k$, we terminate the clustering and obtain the final connected components of $S_{c}$ as clustering results. $\lambda_{max}$ is an empirical parameter that controls the dropping rate of edges and prevents data overflow. Besides, we simplify the distance matrix to accelerate optimization via keeping $r$ nearest neighbors and setting other edges as 0 to reduce the variable scale of each view from $n^{2}$ to $n*r$ ($r<<n$). By focusing on only the most important edges, optimization is accelerated without sacrificing accuracy. The ablation study on the hyper-parameter $r$ and $\lambda_{max}$ are shown in Section~\ref{sec:hyper_analysis}. 

The training procedures of ANN and ANNLD are exhibited in Algorithm \ref{ann algorithm} and Algorithm \ref{annld algorithm}, respectively.

\begin{algorithm}[h]
  \caption{Training procedure of ANN }
  \label{ann algorithm}
 % \scriptsize                                                 
  \begin{algorithmic}[1]
   \Require Target class number $k$, View Structure $V$, Multi-view data $X$
   \State Initialize latent representation $z_{v}$ for each subview
   \State Initialize each subview connection graph $S_{v}$ and consensus connection graph $S_{c}$ by \textbf{Eq. (\ref{agglomaration})} and \textbf{Eq. (\ref{sigma})}
   \State Initialize each subview raw information $D_{v}$ and initialize agglomerated raw information $D_{c}$ by \textbf{Eq. (\ref{dv formula})}
   \State Initialize the eigenvalue matrix $F$ of $S_{c}$
    \While {\textit{cluster number}$<k$}
    \State Fix $F$
    \State Update $\mathcal{L}_{sc},\mathcal{L}_{gc},\mathcal{L}_{cac}$ by \textbf{Eq. (\ref{CLRLOSS})}, \textbf{Eq. (\ref{gc loss})}, and \textbf{Eq. (\ref{ann cac})}
    \State Update $Z^{'},W^{'}\leftarrow$ Adam($\mathcal{L}_{sc},\mathcal{L}_{gc},\mathcal{L}_{cac}$)
    \State Fix $Z^{'},W^{'}$
    \State Update $S_{c}^{'}$ by $Z^{'},W^{'}$ by \textbf{Eq. (\ref{sc formula})}
    \State Update $F^{'}$ by $S_{c}^{'}$
    \If {\textit{cluster number}$>k$}
    \State Resume network parameters
    \State $\lambda = \lambda / 2$
    \Else {\textit{cluster number}$<k$}
    \State $S_{c},Z,W,F\leftarrow S_{c}^{'},Z^{'},W^{'},F^{'}$
    \State $\lambda=\min(\lambda_{max}, 2*\lambda)$
    \EndIf
    \EndWhile
    \State Obtain results by connected components of $S_{c}$
  \end{algorithmic}
 % \vspace{-0.1cm}
\end{algorithm}

\begin{algorithm}[h]
  \caption{Training procedure of ANNLD }
  \label{annld algorithm}
 % \scriptsize                                                 
  \begin{algorithmic}[1]
   \Require Target class number $k$, View Structure $V$, Multi-view data $X$
   \State Initialize $z_{v},S_{v}$ for each subview and $S_{c}$ for consensus view by \textbf{Eq. (\ref{agglomaration})} and \textbf{Eq. (\ref{Learnable b})}
   \State Initialize the eigenvalue matrix $F$ of $S_{c}$
    \While {\textit{cluster number}$<k$}
    \State Update $X^{'}$ by \textbf{Eq. (\ref{Learnable h})}
    \State Use $X^{'}$ to update $D_{v}$ for each subview and initialize agglomerated raw information $D_{c}$ by \textbf{Eq. (\ref{dv formula})}
    \State Fix $F$
    \State Update $\mathcal{L}_{sc},\mathcal{L}_{gc},\mathcal{L}_{cac}$ by \textbf{Eq. (\ref{CLRLOSS})}, \textbf{Eq. (\ref{gc loss})}, and \textbf{Eq. (\ref{Learnable b 2})}
    \State Update $Z^{'},W^{'},H^{'},B^{'}\leftarrow$ Adam($\mathcal{L}_{sc},\mathcal{L}_{gc},\mathcal{L}_{cac}$)
    \State Fix $Z^{'},W^{'},H^{'},B^{'}$
    \State Update $S_{c}^{'}$ by $Z^{'},W^{'},H^{'},B^{'}$ by \textbf{Eq. (\ref{Learnable b})}
    \State Update $F^{'}$ by $S_{c}^{'}$
    \If {\textit{cluster number}$>k$}
    \State Resume network parameters
    \State $\lambda = \lambda / 2$
    \Else 
    \State $S_{c},Z,W,H,B,F\leftarrow S_{c}^{'},Z^{'},W^{'},H^{'},B^{'},F^{'}$
    \State $\lambda=\min(\lambda_{max}, 2*\lambda)$
    \EndIf
    \EndWhile
    \State Obtain results by connected components of $S_{c}$
  \end{algorithmic}
 % \vspace{-0.1cm}
\end{algorithm}
%can effectively embrace these terms and data-driven extension in a unified neural network and is convenient to train. 
\section{Experiments}\label{expe}
\subsection{Experimental Setup}
\begin{table*}[!htb]
    \caption{Best Clustering Performance over Four Public Datasets}
    \label{main results}
    \small
    \centering
    \begin{tabular}{c|cccccc||cccccc}
    \toprule
 & \multicolumn{6}{c|}{HW} & \multicolumn{6}{c}{MNIST-USP} \\
 \midrule
Method & NMI & RI & Purity & Precision & Recall & F-Score & NMI & RI & Purity & Precision & Recall & F-Score \\
\midrule
SC &  0.591&  0.886&  0.687&  0.443&  0.568&  0.497&  0.653&  0.922&  0.733&  0.604&  0.620&  0.612\\
Co-reg &  0.761&  0.943&  0.835&  0.703&  0.741&  0.721&  0.755&  0.738&  0.826&  0.715&  0.738&  0.726\\
Co-training &  0.775&  0.946&  0.841&  0.723&  0.751&  0.736&  0.829&  0.964&  0.903&  0.811&  0.835&  0.823\\
SWMC &  0.946&  0.990&  0.975&  0.950&  0.951&  0.951&  0.860&  0.941&  0.897&  0.655&  0.864&  0.745\\
MVGL &  0.885&  0.974&  0.936&  0.860&  0.881&  0.870&  0.690&  0.892&  0.752&  0.470&  0.673&  0.553\\
BMVC &  0.715&  0.903&  0.805&  0.508&  0.720&  0.596&  0.481&  0.866&  0.598&  0.367&  0.476&  0.414\\
MLAN &  0.938&  0.989&  0.973&  0.945&  0.946&  0.946&  0.871&  0.952&  0.901&  0.713&  0.869&  0.784\\
GMC &  0.904&  0.972&  0.949&  0.826&  0.908&  0.865&  \textbf{0.994} &  \textbf{0.999}&  \textbf{0.998}&  \textbf{0.996}&  \textbf{0.996}&  \textbf{0.996}\\
LT-MSC &  0.855&  0.970&  0.920&  0.850&  0.853&  0.851&  0.719&  0.928&  0.790&  0.624&  0.696&  0.658\\
COMIC &  0.886&  0.976&  0.936&  0.877&  0.883&  0.880&  0.757&  0.870&  0.933&  0.427&  0.906&  0.581\\
\midrule
ANN &  \textbf{0.951}&  \textbf{0.992}&  \textbf{0.979}&  \textbf{0.958}&  \textbf{0.959}&  \textbf{0.958}&  0.866&  0.954&  0.906&  0.732&  0.857&  0.790\\
\midrule
 & \multicolumn{6}{c|}{ALOI}& \multicolumn{6}{c}{Caltech}\\
 \midrule
Method & NMI & RI & Purity & Precision & Recall & F-Score & NMI & RI & Purity & Precision & Recall & F-Score \\
\midrule
SC &  0.776&  0.928&  0.827&  0.698&  0.744&  0.720&  0.423&  0.771&  0.335&  0.671&  0.193&  0.300\\
Co-reg &  0.669&  0.890&  0.772&  0.551&  0.670&  0.603&  0.611&  0.799&  0.442&  0.872&  0.246&  0.384\\
Co-training &  0.722&  0.913&  0.821&  0.633&  0.719&  0.673&  0.654&  0.806&  0.437&  0.921&  0.258&  0.402\\
SWMC &  0.880&  0.903&  0.989&  0.563&  0.979&  0.715&  0.654&  0.762&  0.716&  0.531&  0.540&  0.536\\
MVGL &  0.844&  0.942&  0.883&  0.735&  0.828&  0.779&  0.637&  0.787&  0.585&  0.622&  0.412&  0.495\\
BMVC &  0.637&  0.850&  0.709&  0.429&  0.634&  0.512&  0.595&  0.800&  0.427&  0.860&  0.255&  0.393\\
MLAN &  0.860&  0.943&  0.835&  0.737&  0.737&  0.783&  0.787&  0.884&  0.778&  0.896&  0.613&  0.728\\
GMC &  0.794&  0.883&  0.827&  0.518&  0.790&  0.626&  0.764&  0.878&  0.777&  0.866&  0.612&  0.717\\
LT-MSC &  0.846&  0.953&  0.900&  0.792&  0.843&  0.817&  0.664&  0.816&  0.523&  \textbf{0.928}&  0.298&  0.451\\
COMIC &  0.599&  0.865&  0.761&  0.466&  0.631&  0.536&  0.670&  0.815&  0.498&  0.911&  0.302&  0.454\\
\midrule
ANN &  \textbf{0.980}&  \textbf{0.996}&  \textbf{0.991}&  \textbf{0.982}&  \textbf{0.982}&  \textbf{0.982}&  \textbf{0.829}&  \textbf{0.927}&  \textbf{0.896}&  0.898&  \textbf{0.636}&  \textbf{0.745}\\
\bottomrule
\end{tabular}
\end{table*}
We compare our models with a number of state-of-the-art algorithms: Spectral Clustering (SC) \cite{scikit-learn}, Co-trained Spectral Clustering (Co-training) \cite{cotraining2011}, Co-regularized Spectral Clustering (Co-reg) \cite{coreg2011}, Binary Multi-view clustering (BMVC) \cite{bmvc}, Graph Learning for Multi-view Clustering (MVGL) \cite{mvgl}, Self-weighted Multi-view Clustering (SWMC) \cite{swmc}, Multi-view Learning with Adaptive neighbors (MLAN) \cite{aaai2017}, Low-rank Tensor constrained Multi-view Subspace Clustering (LT-MSC) \cite{zhang2015low}, Graph-based Multi-view Clustering (GMC) \cite{wang2019gmc}, and Cross-view Matching Clustering (COMIC) \cite{peng2019comic}. We apply $K$-means to help COMIC get exact clusters.
We evaluate them on four widely-used multi-view datasets and one dataset prepared by ourselves:
%against eight state-of-the-art multi-view clustering algorithms
%The four conventional multi-view datasets exhibit the promising ability of ANN in analyzing multi-view datasets, and \textcolor{red}{the case study} presents the advances and the extensibility of the agglomerative method in complex view structures.
%We run experiments on four public image datasets: two object recognition datasets and two handwritten digit datasets. We further take a private survey dataset as the case study. We construct the datasets as below.

\textbf{UCI Handwritten numerals (HW)}~\cite{baruah2015dataset} consists of 2,000 sample, 200 records of digit 0 to 9 respectively. We use the six public descriptor features provided by the data for training: 76-dimension Fourier coefficients of the character shape features, 216-dimension profile correlation features, 64-dimension Karhunen-love coefficient features, 240-dimension pixel average features in 2 $\times$ 3 windows, 47-dimension Zernike moment features, and 6-dimension morphological features.

\textbf{MNIST-USPS} dataset comprises two commonly used handing written digit datasets: MNIST~\cite{lecun1998gradient} and USPS~\cite{demiriz2002linear}. We randomly pick 400 samples from 10 digits and consider two datasets as two independent views. The constructed dataset is composed of 4000 samples with 784 dimensions for MNIST and 256 dimensions for USPS.

\textbf{Amsterdam Library of Object Images (ALOI)}~\cite{geusebroek2005amsterdam} picks all 879 images of 8 objects (Object Number: 65, 121, 138, 262, 583, 783, 822, and 868). Four public descriptor features are used: first 13-dimension Haralick features (radius 1 pixel), 216-dimension RGB color histogram features, 27-dimension Hue-Saturation-Brightness color histogram features, and 77-dimension color similarity features.

\textbf{Caltech101}~\cite{fei2004learning} contains 2,386 images from 16 classes. Following the setting in previous work\cite{fanello2014dictionary}, we keep the samples that share the same class with 5 of 10 most similar neighbors. The dataset embraces six diverse views: 48-dim Gabor feature, 40-dim wavelet moments (WM), 254-dim CENTRIST feature, 1,984-dim HOG feature, 512-dim GIST feature, and 928-dim LBP feature.

\textbf{Survey} is provided by a local financial company. It exhibits a complex data structure and contains consumers' investment risk preferences assessed at six levels based on 71 reliable consumers' investigation feedback.
%collected by a local Micro-invest company.
It has a two-layer view structure with 75 dimensions. These dimensions can be divided into 11 independent views consisting various questions, e.g., \textit{concerns to environment} and \textit{advance spirit in life} in $V^{(0)}$; the aspects can be further sorted into two general attitudes based on domain knowledge~\cite{hartmann2012consumer,mandrik2005exploring}, i.e., the domain views in the first layer $V^{(1)}$: Company Social Responsibility (CSR) and Emotion and Advance Rating (EAR). The consensus view $S^{(c)}$ in the second layer $V^{(2)}$ will be the fusion of CSR and EAR.

%, we apply $K$-means to assist further clustering. 
%We run experiments on five datasets below. 
%We evaluate the best clustering performance of the algorithms using five criteria: Normalized Mutual Information (NMI), Purity, Precision, Recall, and F-Score. Specially, we replace Purity by Rand Index (RI), on the Survey dataset, with is small and imbalanced distributed.

%To specify the details of experiment,
%In our experiments, 
We initialize $\lambda=15$, $D$ by the L2 norm matrix of raw data and select the nearest 10 and 9 neighbors of distance matrices to execute the clustering for ANN and ANNLD, respectively. $\forall h_{j}\in H,\forall b^{(v)}\in B, \forall w^{(v)}\in W$, we set $h_{j}=1, b^{(v)}=0, w^{(i)}=1/|v|$. The same setting is used for ANN over four datasets: $\lambda_{max}=10^{5},\mathcal{P}=1.13,lr=0.05$, where $\lambda_{max}$ is the maximum of $\lambda$ to prevent data overflow during optimization. Similarly, we set ANNLD: $\lambda_{max}=10^{7},\mathcal{P}=1.05,lr=0.1$.
We evaluate the algorithms using six criteria: Normalized Mutual Information (NMI), Rand Index (RI), Purity, Precision, Recall, and F-Score.
%The precision, recall and F-score compare every two points to measure the validity of pairwise points' cluster relationship.
We do not use Purity to evaluate Survey, because Purity may be invalid if the dataset is imbalanced distributed.

\subsection{Experiment Performance}\label{Result}
\begin{table}
    \small
    \caption{Best Clustering Performance on Survey Dataset}
    \label{survey results}
    \centering
    \begin{tabular}{c|ccccc}
    \toprule
    Method  & NMI  & RI & Precision & Recall & F-score\\
    \midrule
    SC & 0.126& 0.326& 0.253& \textbf{0.813}& 0.386\\
Co-reg & 0.171& 0.665& 0.286& 0.190& 0.228\\
Co-training & 0.185& 0.678& 0.315& 0.202& 0.246\\
SWMC & 0.146& 0.443& 0.229& 0.481& 0.310\\
MVGL & 0.145& 0.457& 0.284& 0.714& 0.407\\
BMVC & 0.124& 0.632& 0.283& 0.269& 0.276\\
MLAN & 0.158& 0.567& 0.256& 0.348& 0.295\\
GMC & 0.101& 0.491& 0.251& 0.479& 0.329\\
LT-MSC & 0.150& 0.671& 0.293& 0.185& 0.227\\
COMIC & 0.142& 0.659& 0.296& 0.224& 0.255\\
\midrule
ANN & 0.178& 0.580& 0.256& 0.321& 0.285\\
ANNLD & \textbf{0.262}& \textbf{0.686}& \textbf{0.351}& 0.759& \textbf{0.480}\\
    \bottomrule
    \end{tabular}
\end{table}

\begin{figure}
    \centering
    \includegraphics[width=0.7\linewidth]{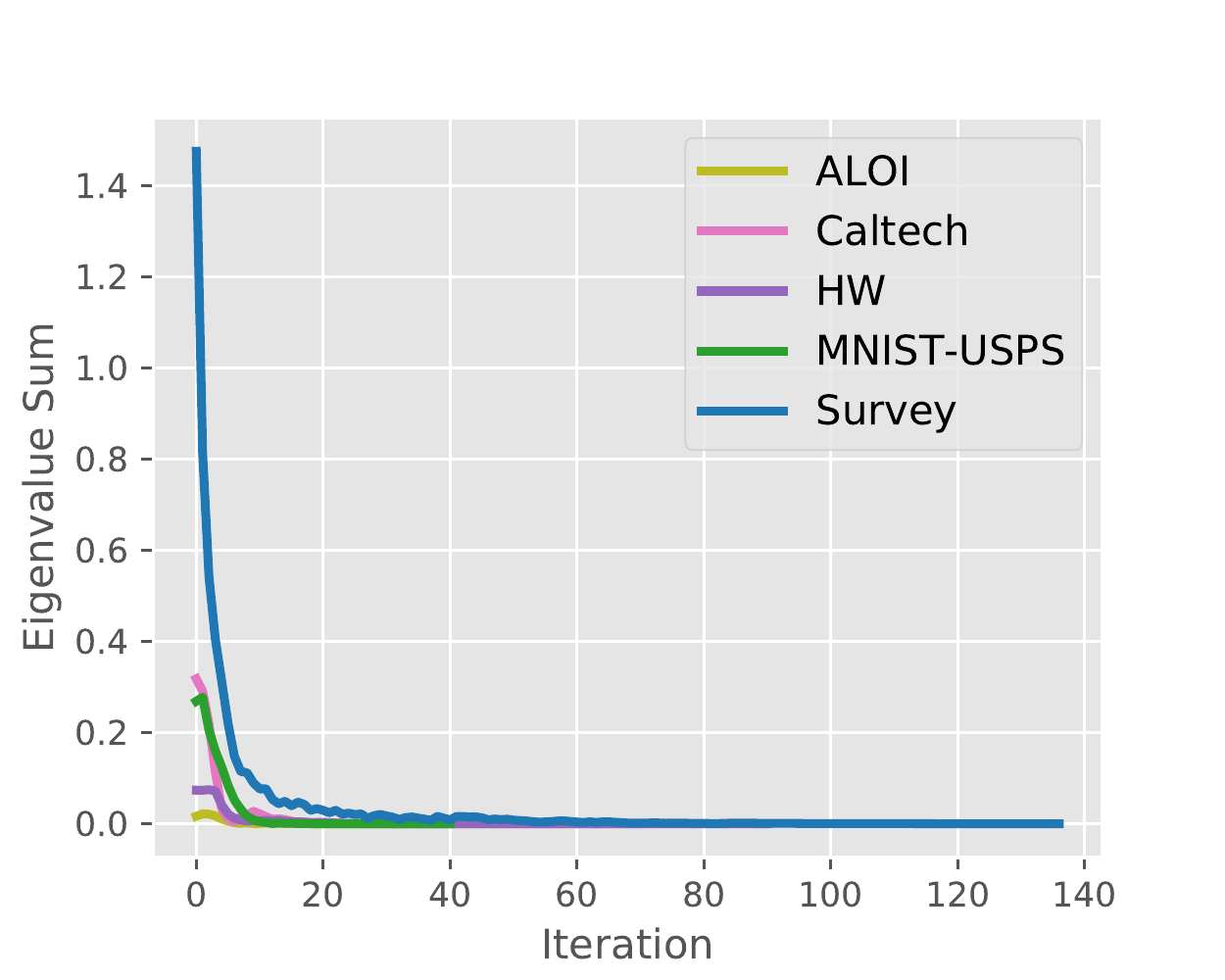}
    \caption{
    %Constrained Laplacian rank convergence analysis on $k$-partition clustering over five datasets. Y-axis represents the sum of $k$ smallest eigenvalues.
    The sum of $k$ smallest eigenvalues on $k$-partition clustering over five datasets.
    }
    \label{Eigenvalue loss}
\end{figure}

Table \ref{main results} reveals the promising ability of ANN in analyzing multi-view data. We can observe that ANN outperforms all state-of-the-art on ALOI and Caltech. In particular, ANN improves in NMI, Purity, and F-Score by 0.063, 0.115, 0.115 on ALOI and 0.042, 0.118, 0.017 on Caltech. ANN also achieves the state-of-the-art performance on HW and MNIST-USP. Although GMC achieves the best performance on MNIST-USP, it shows bad robustness on other datasets, e.g., from the aspect of NMI on ALOI and Caltech, GMC only obtains 0.794 and 0.764 on ALOI while ANN achieves 0.980 and 0.829, respectively. 
Table \ref{survey results} shows the methods' performance on the multi-layer Survey dataset. %Since Survey has a two-layer view structure, 
Most of the multi-view algorithms, e.g., GMC and BMVC, cannot work well with the two-layer view structure of Survey that their NMI scores are lower than SC's. Both ANN and ANNLD achieve excellent performance on Survey. We can observe that ANN without learnable data space can still achieve the state-of-the-art performance and ANNLD improves NMI and F-Score by 0.077 and 0.073 than Co-training, demonstrating the effectiveness of proposed agglomerative analysis in utilizing complex view structure. ANNLD obtains better performance on all the matrices than ANN, which shows the effectiveness of learnable data space. Note that although SC achieves better Recall performance than ANN and ANNLD, it does not mean that SC is superior to the proposed methods, because SC's best performance is achieved by partitioning almost all samples to a single cluster.

In all, our proposed ANN and ANNLD consistently show the best robustness and performance over five diverse datasets. The agglomerative consensus analysis and learnable data space can enhance the methods' ability to analyze standard multi-view datasets, as well as handling the multi-layer structured dataset effectively. 

\begin{figure}
    \centering % <-- added
    \begin{minipage}{0.42\textwidth}
    \centerline{\includegraphics[width=1\textwidth]{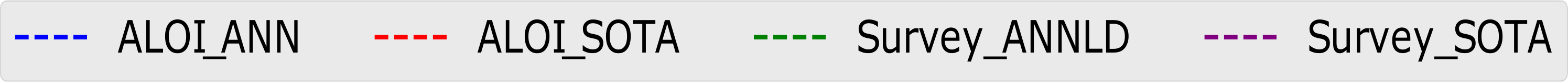}}
    \vspace{-2mm}
    \end{minipage}
    \subfigure[Exponent of $\lambda_{max}$ ]{\includegraphics[width=0.23\textwidth]{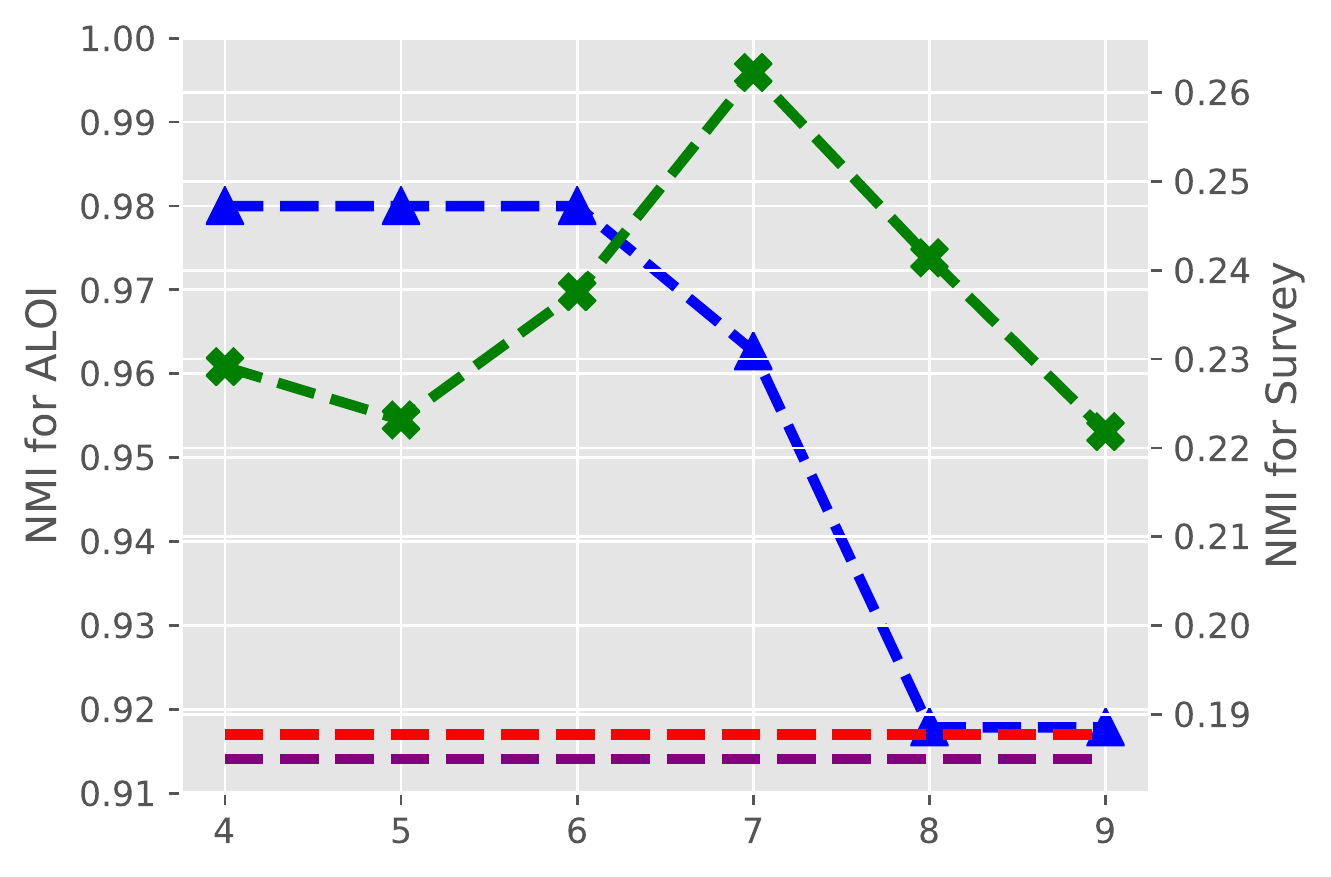}}
    \subfigure[Parameter-$lr$]{\includegraphics[width=0.23\textwidth]{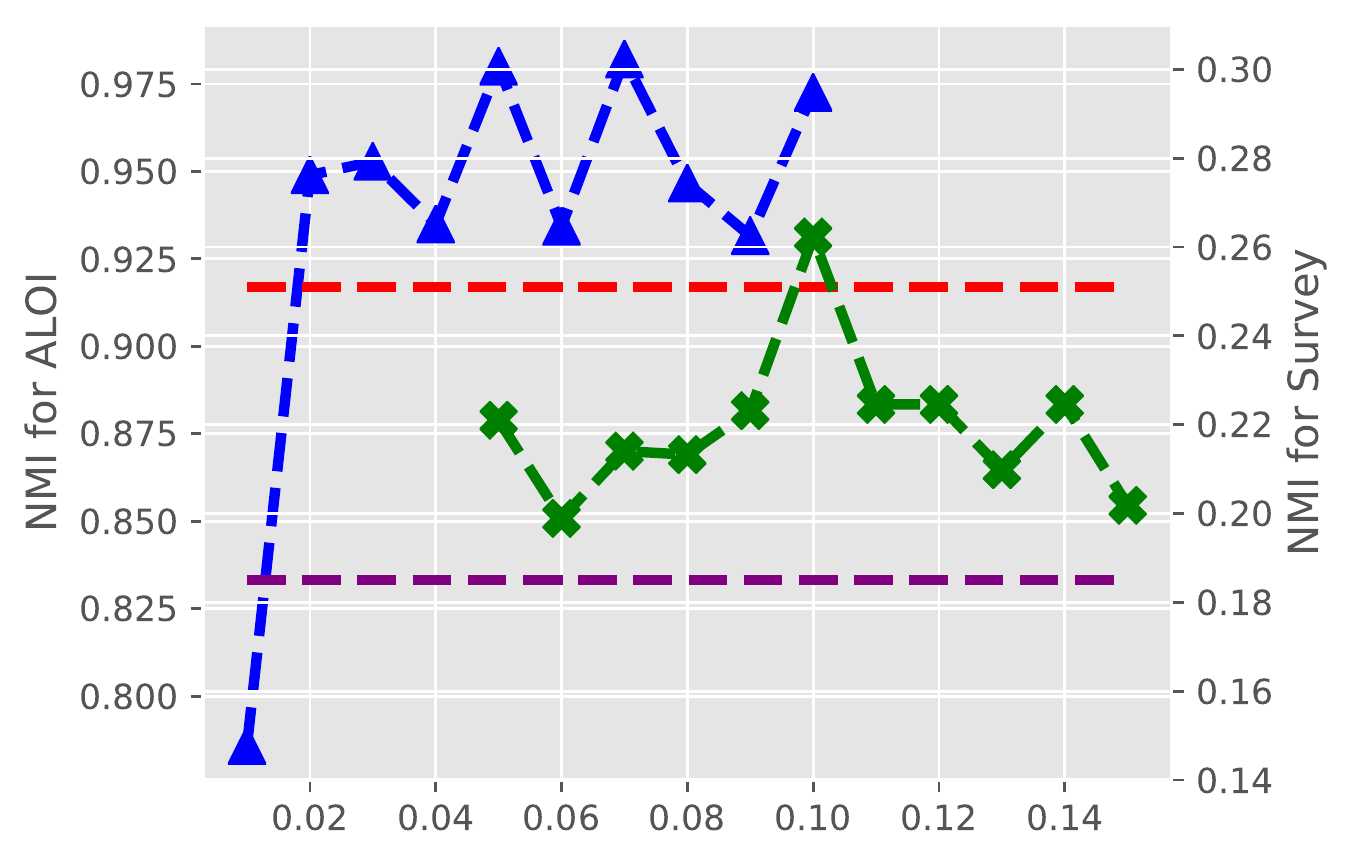}}
    \\
    \subfigure[neighbor Number]{\includegraphics[width=0.23\textwidth]{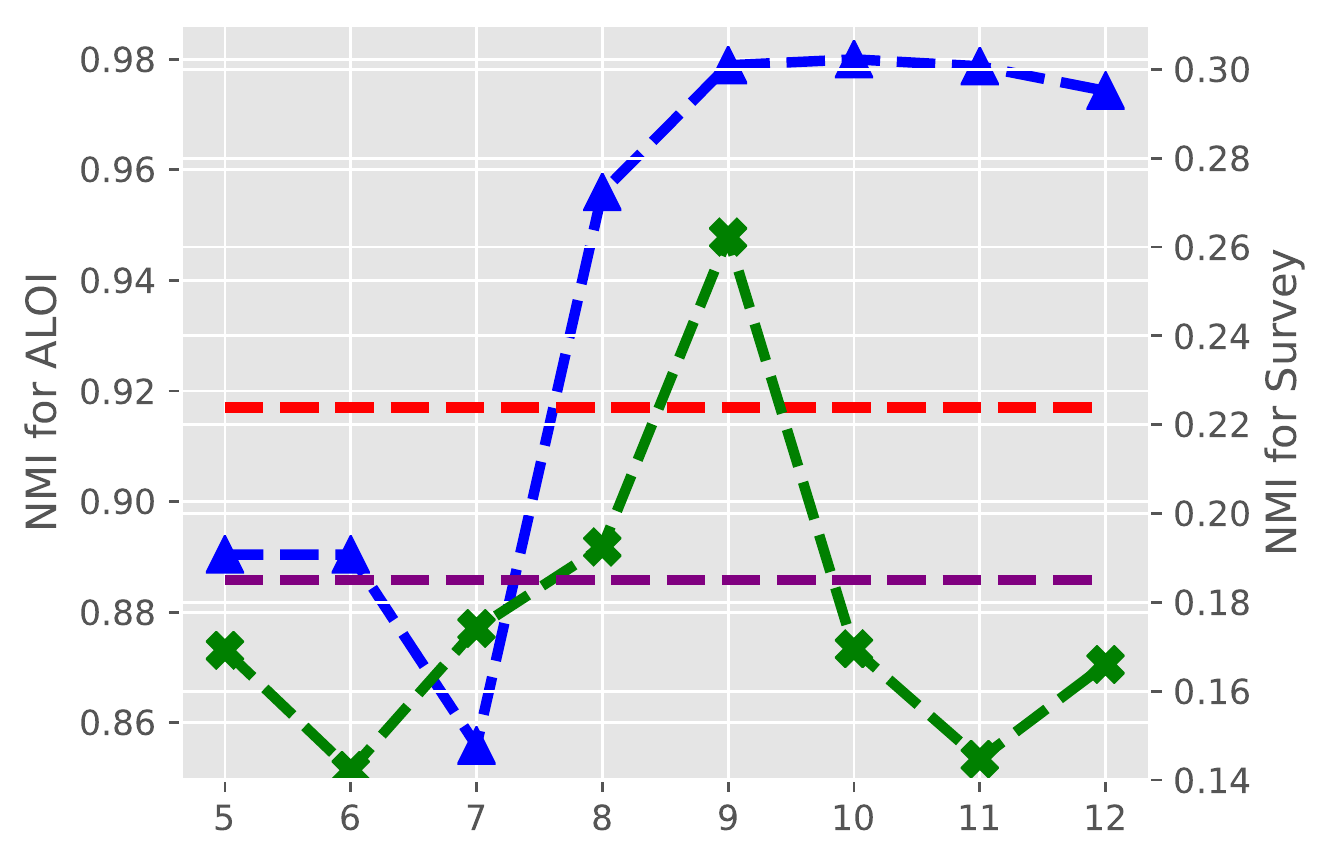}}
    \subfigure[Parameter-$\mathcal{P}$]{\includegraphics[width=0.23\textwidth]{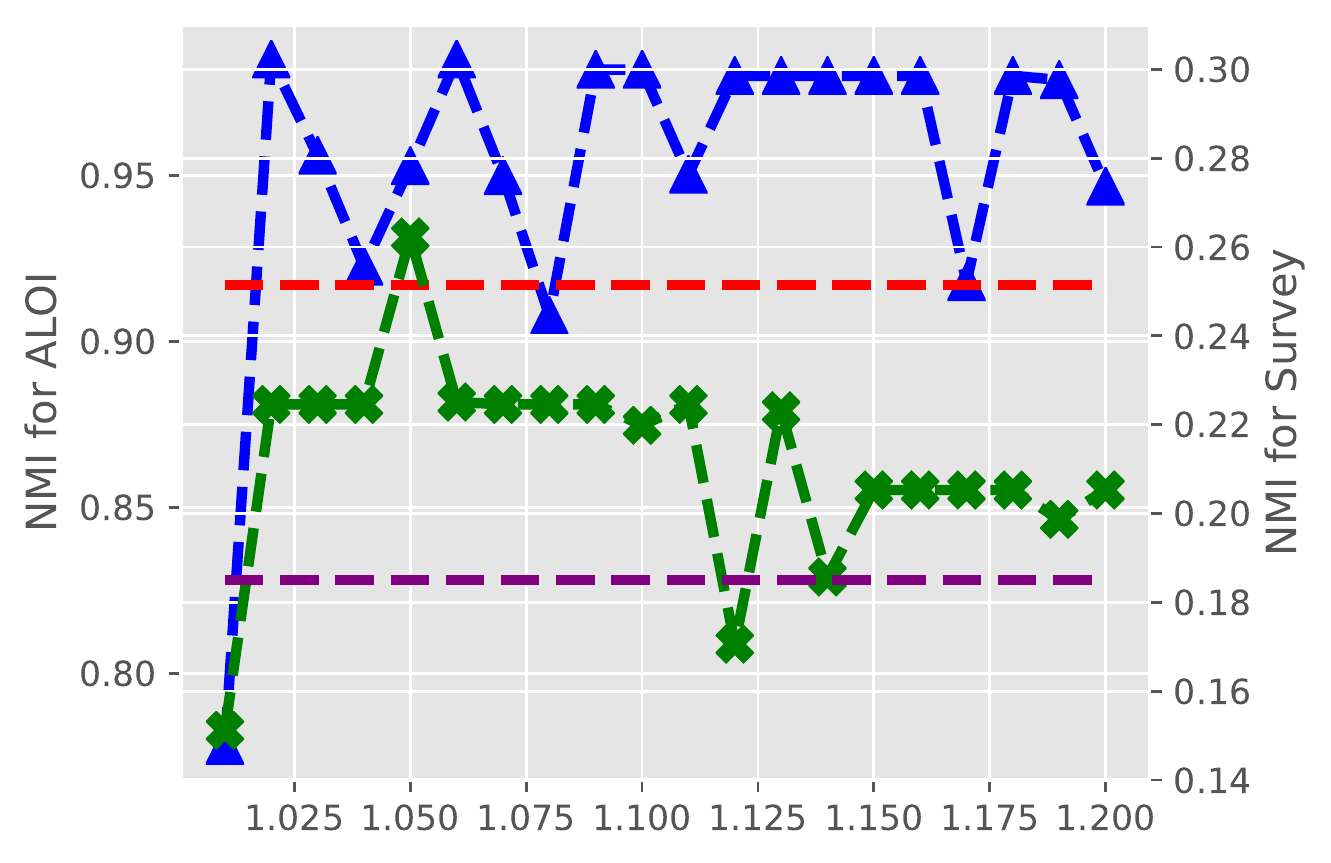}}
    \caption{Our model's NMI under varying hyper-parameters on ALOI and Survey.}
    \label{hyper}
\end{figure}
    
\begin{figure*}
    \centering % <-- added
    \subfigure[$t=0$ (Raw Data)]{\includegraphics[width=0.23\textwidth]{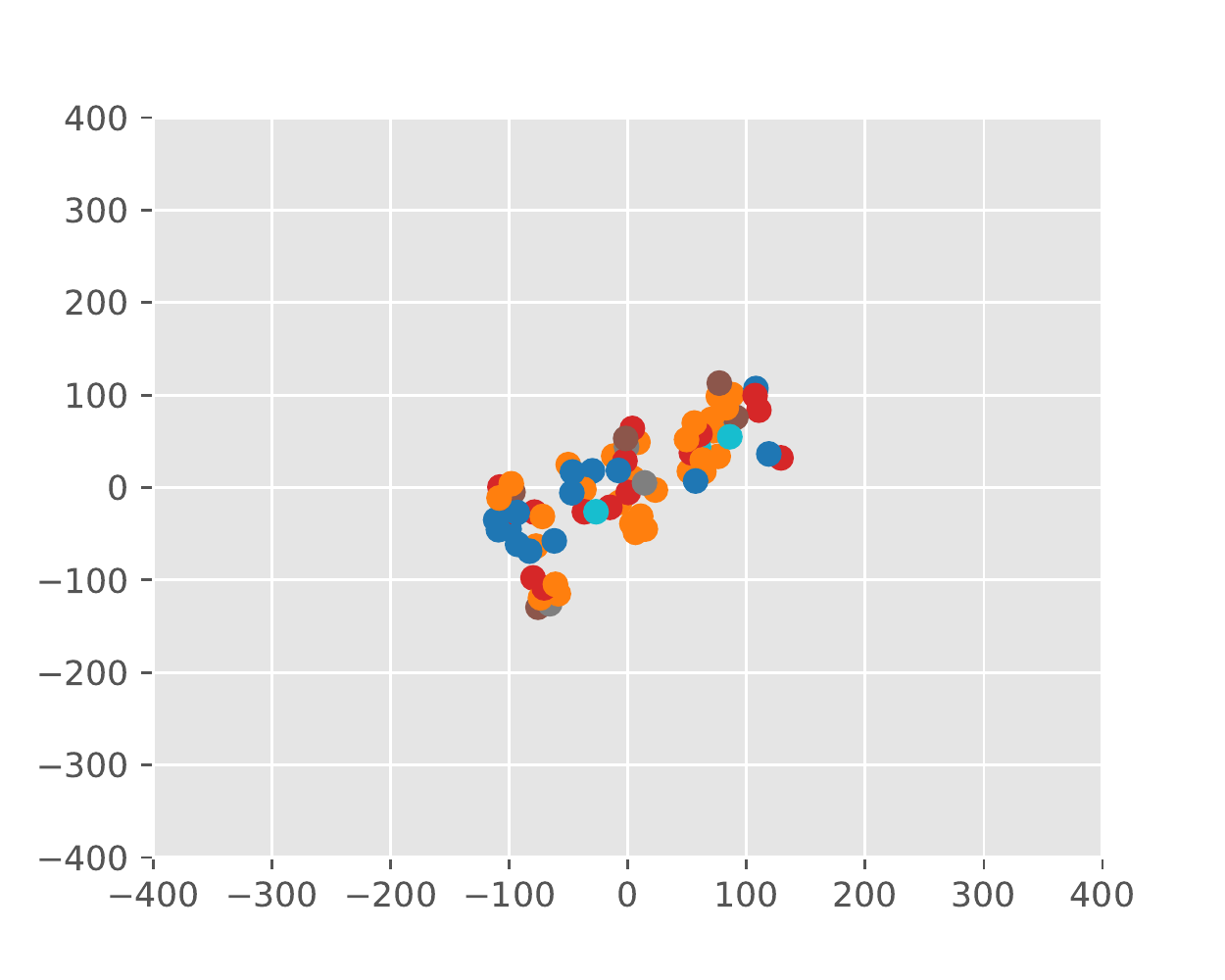}}
    \subfigure[$t=10$]{\includegraphics[width=0.23\textwidth]{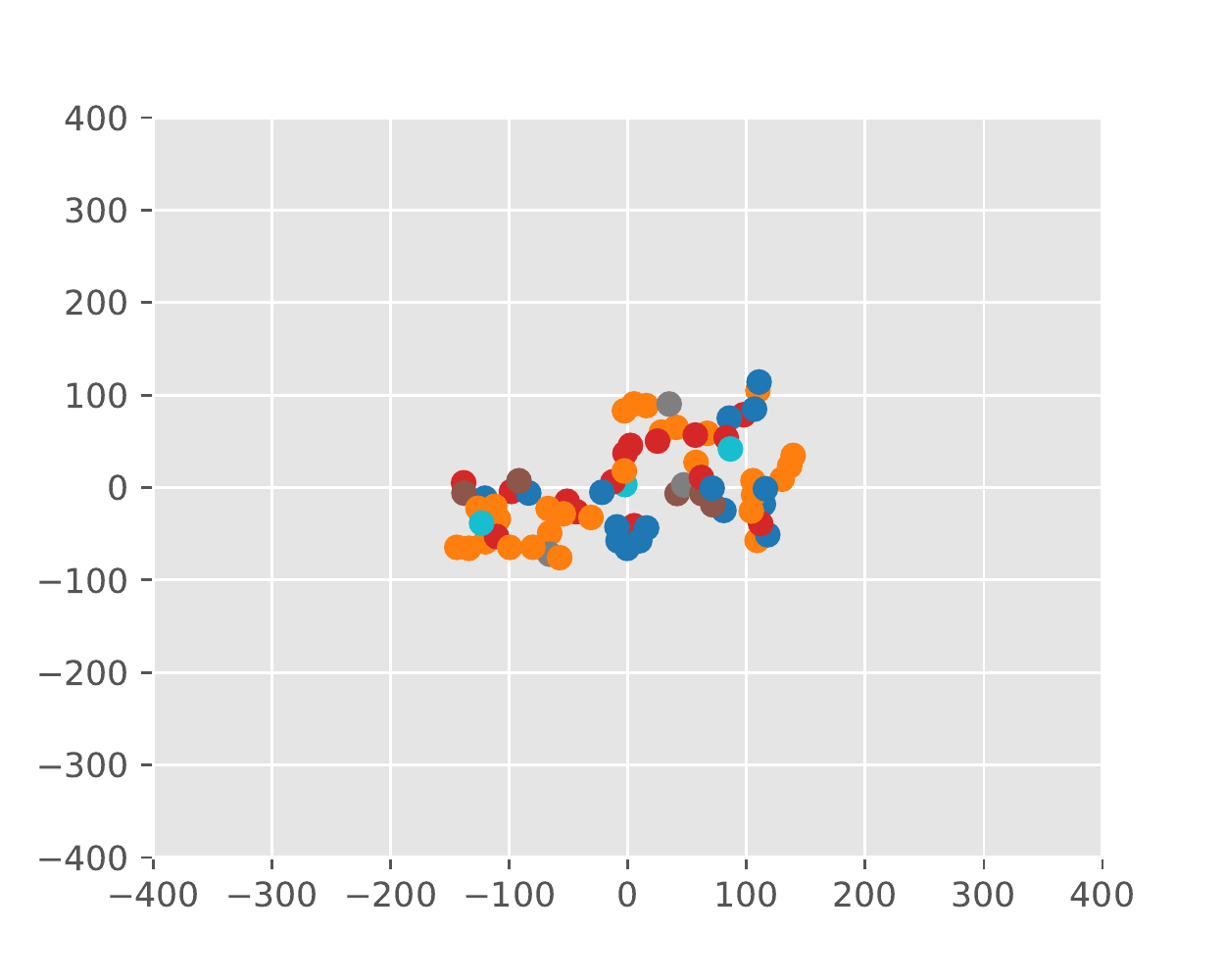}}\vspace{-2mm}
    \subfigure[$t=30$]{\includegraphics[width=0.23\textwidth]{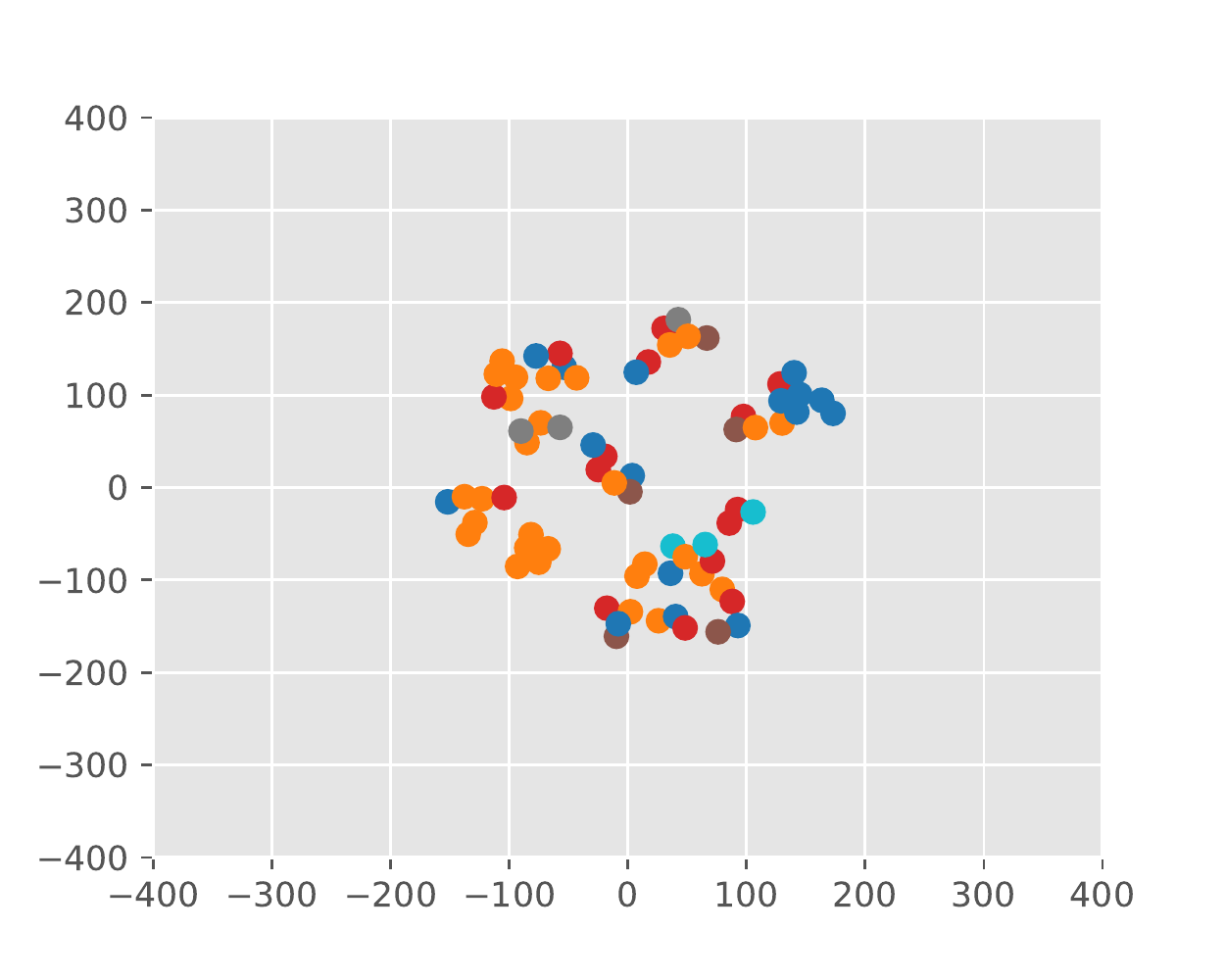}}
    \subfigure[$t=84$ (Converged)]{\includegraphics[width=0.23\textwidth]{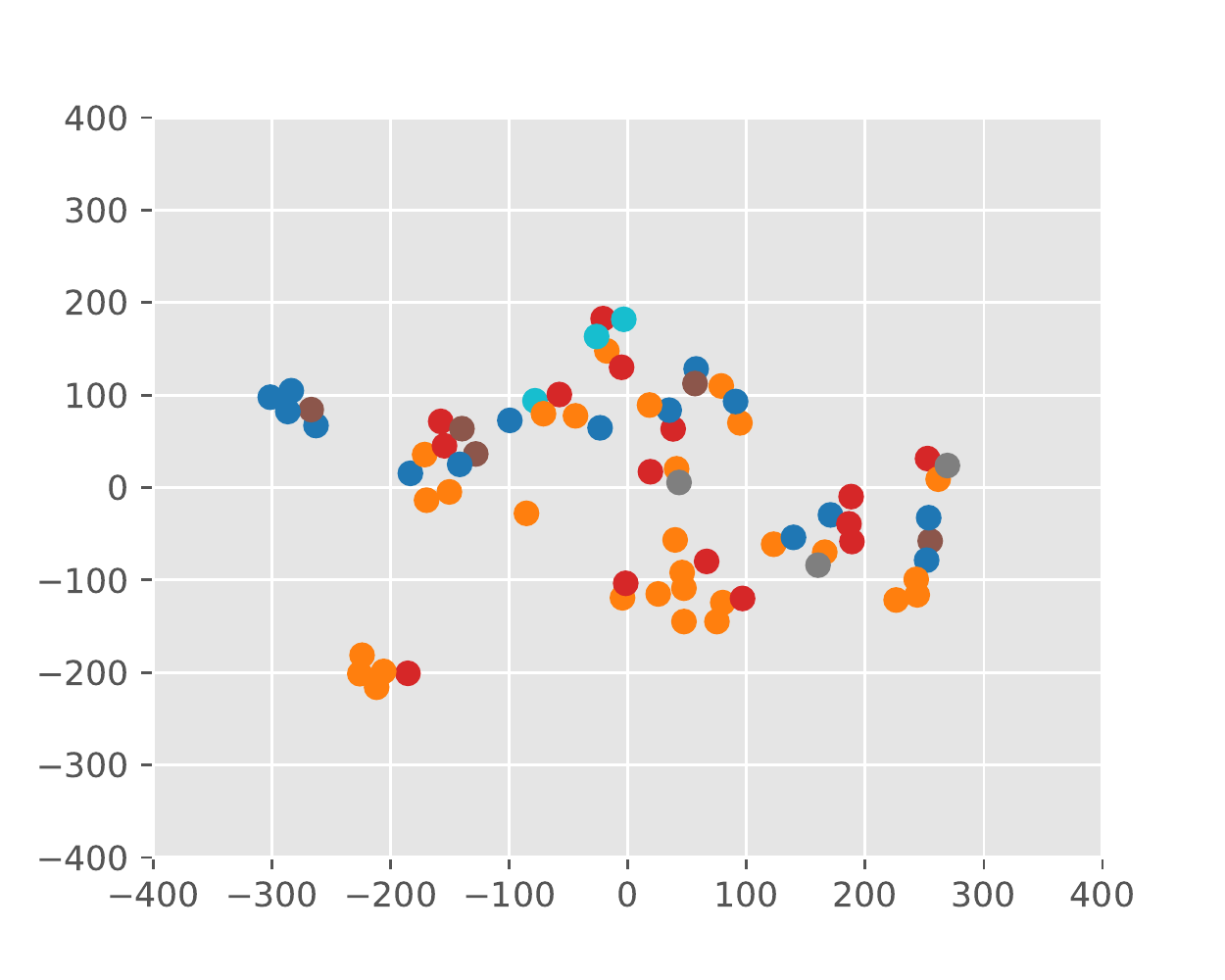}}
\caption{Visualization of Two-layered Survey data embedding space.}
\label{embedding}
\end{figure*}
\begin{figure*}
    \centering % <-- added
    \subfigure[$t=0$ (Raw Graph)]{\includegraphics[width=0.23\textwidth]{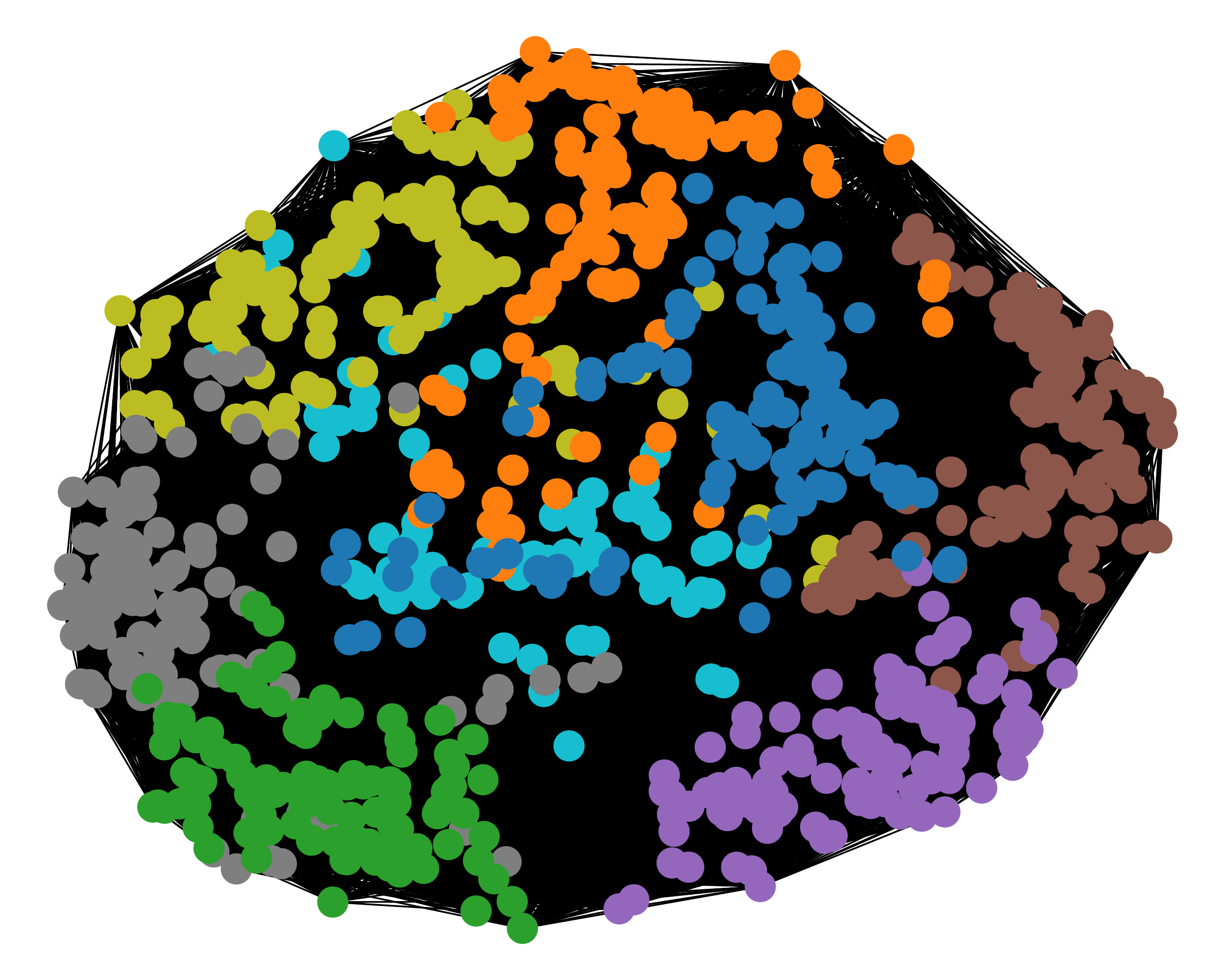}}
    \subfigure[$t=10$]{\includegraphics[width=0.23\textwidth]{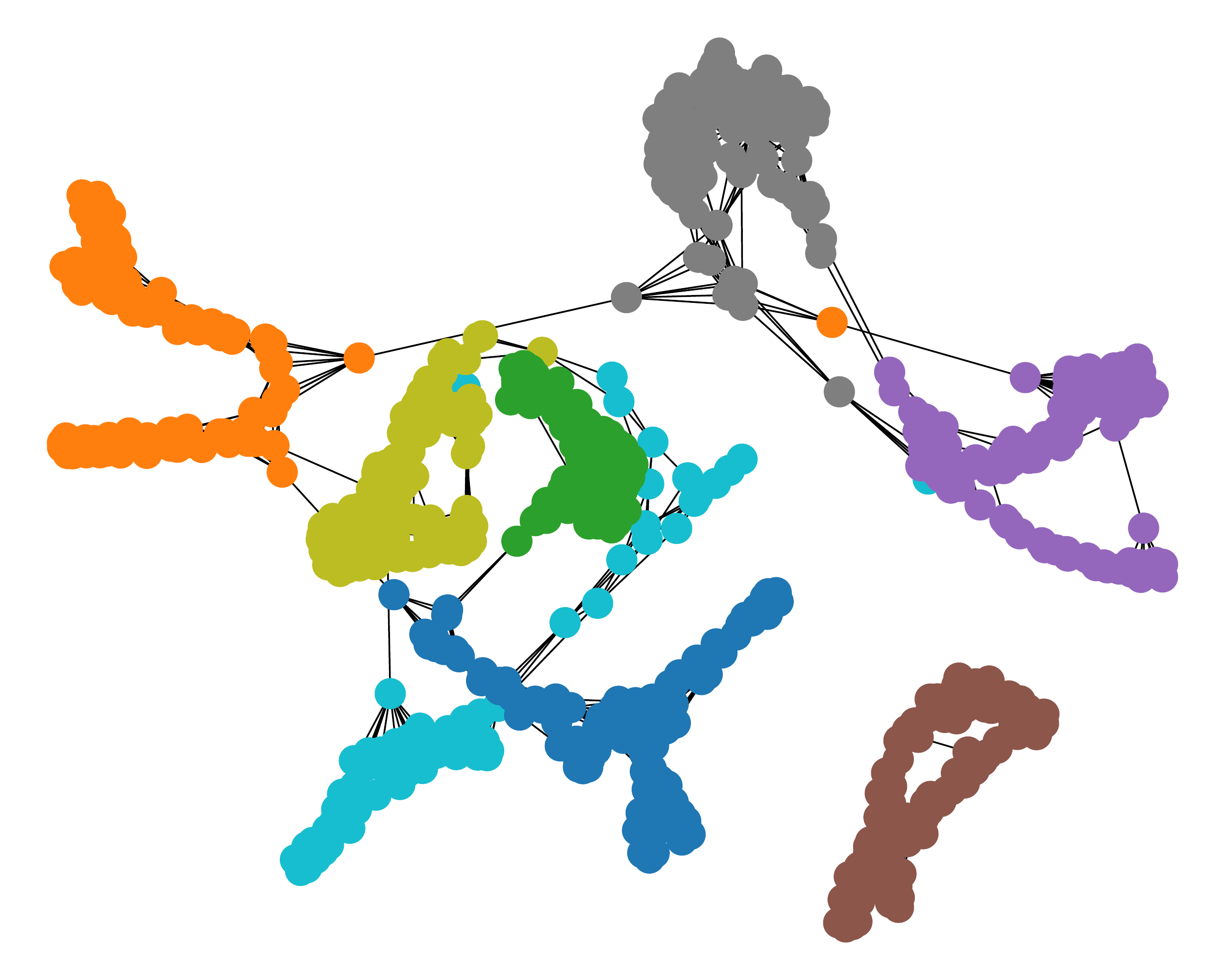}}\vspace{-2mm}
    \subfigure[$t=25$]{\includegraphics[width=0.23\textwidth]{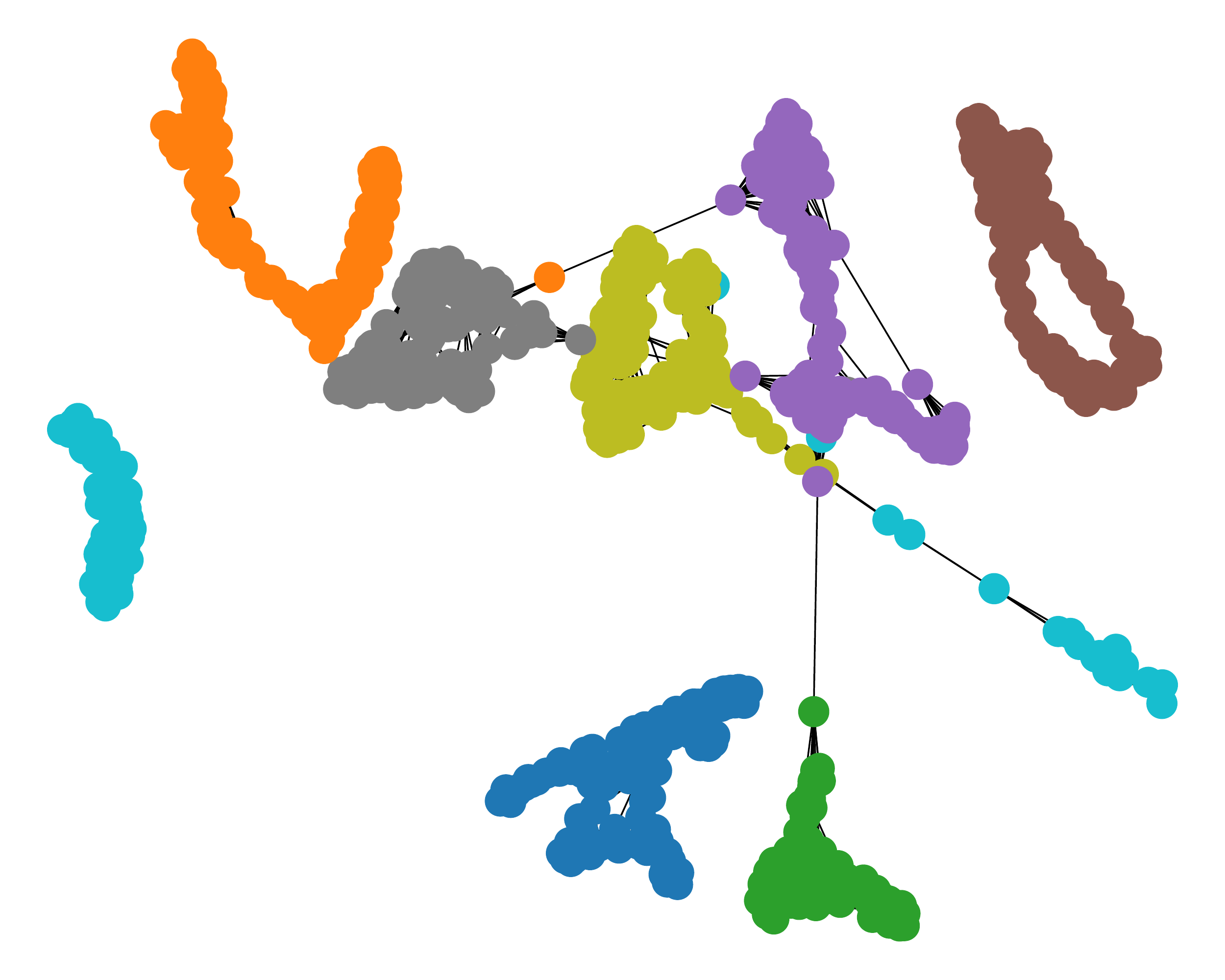}}
    \subfigure[$t=36$ (Converged)]{\includegraphics[width=0.23\textwidth]{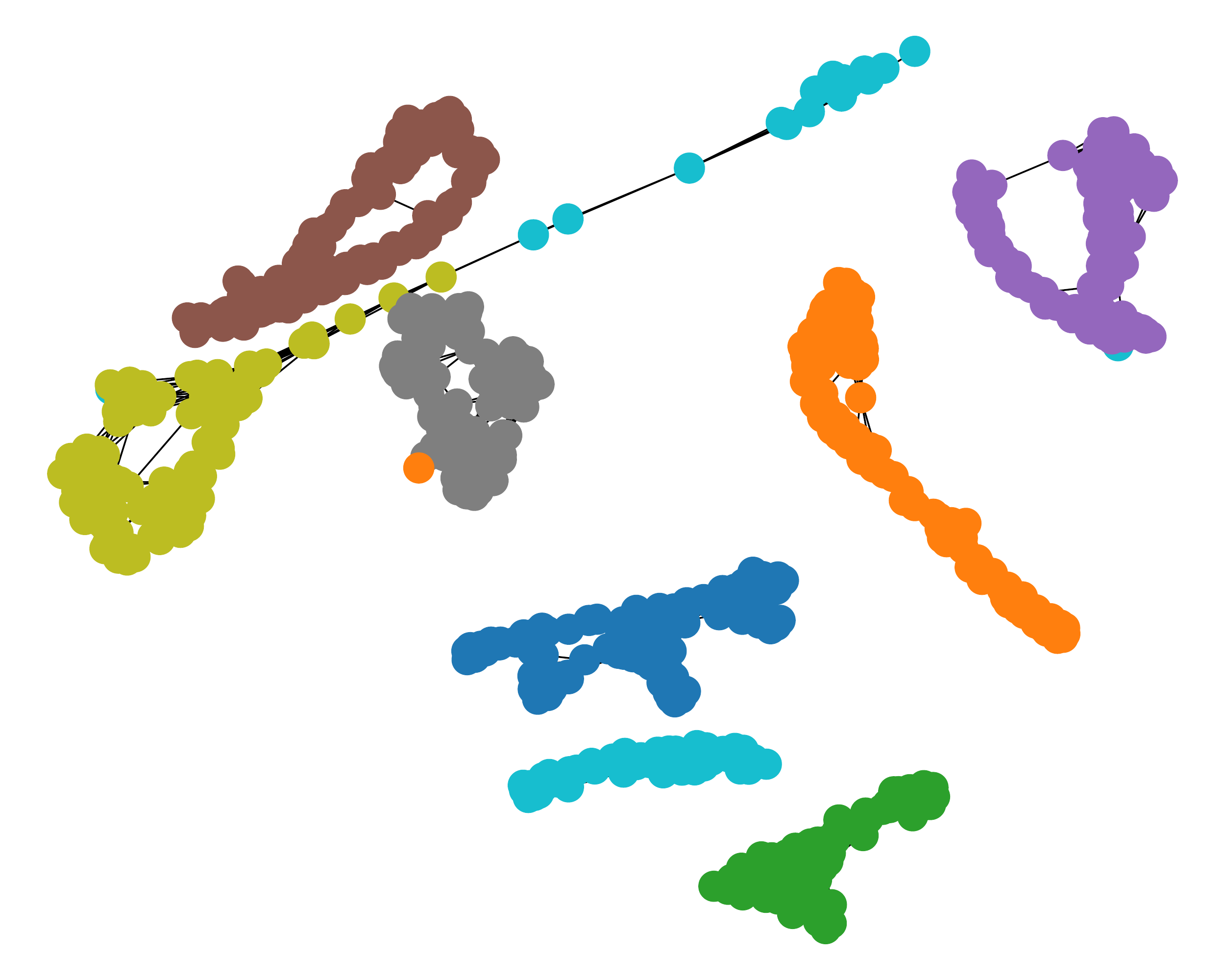}}
\caption{Visualization of ALOI connection graphs.}
\label{connectionn graphs}
\end{figure*}
\subsection{Hyper-parameter and Convergence Analysis}
\label{sec:hyper_analysis}

%\subsection{Hyper-parameter Analysis}\label{hyper_analysis}
We take ALOI and Survey to study the influence of hyper-parameters on ANN and ANNLD, respectively.
We set the learning rate, $lr$, around its optimal values: 0.05 and 0.1, separately; the results (Fig.~\ref{hyper}) show
%, where
%we take ALOI and Survey as an example to show the model's robustness.
%different learning rates around the optimal solution
%separately
%The optimal $lr$ for ANN and ANNLD should be 0.05 and 0.1, so . 
our model is robust in terms of achieving the best performance under most hyper-parameter configurations when compared with the best performance of state-of-the-art (denoted by the red and purple horizontal lines).
%represent the best performance of state-of-the-art, and . 
%However, 
Our networks are predominantly influenced by the number of neighbors, especially on Survey data. In particular, ANNLD may perform poorly if the neighbor number is excessively small or large.
%while the other three parameters only slightly affect the performance. 

Our convergence analysis of the Constrained Laplacian rank of ANN (Fig.~\ref{Eigenvalue loss}) shows that ANN can converge quickly by decreasing the eigenvalue sum to around zero within 20 iterations
%the curve of k smallest eigenvalues during the optimization on five datasets.
%It is evident that t
%The loss function significantly decreases the eigenvalues after iterations. 
%If the initial sum is large, e.g., Survey dataset, the loss function quickly cut meaningless edges to reduce the Laplacian matrix rank and then smoothly cut the remaining edges until convergence.
on all the five datasets, which contain varying quantities from 71 to 4,000.  
%when optimizing the connection graph.
%to decrease the eigenvalue sum near zero in 20 iterations. In conclusion, our method can converge quickly in diverse scenarios.

\subsection{Embedding Visualization of Hidden Data Space}\label{DataSpace}
The embedding graphs indicate the effectiveness of data space projection of ANNLD in Fig.~\ref{embedding}. We apply $K$-Means to transform input data into a 6-dimension distance vector and exhibit the T-distributed Stochastic Neighbor Embedding (t-SNE) with 2-component and 8-perplexity. We can observe that the raw data space has limited capability of distinguishing data that contain overlapping points. After iterations, the embedding points turns to be better distributed and distinct. It is easy to distinguish the purple and green points from other points when the algorithm converges.

\subsection{Visualization of Connection Graphs}
We take ALOI as an example to illustrate how the optimization changes connection graphs in Fig. \ref{connectionn graphs}. We can observe that the raw graph only contains one cluster, and many connections exist between different classes. After iterations, ANN could effectively delete most redundant edges within several iterations. When the optimization converges, ANN can obtain a connection graph with exactly eight connected components. Each component could represent one cluster so that the connected component results could be directly used as the clustering results. Only a few points are clustered into the wrong connected components.
\section{Conclusion}
We propose agglomerative consensus analysis for multi-view clustering.
%with subviews.
%and it shows better performance in ordinary view structures without subviews. 
To this end, we present an extensible Agglomerative Neural Network (ANN) and conduct comprehensive experiments over four public datasets.
We further 
%show the extensibility of ANN, we further
%make data-specific modifications and
%we further make data-specific modifications according to a case study.We 
propose ANN with Learnable Data space (ANNLD) via an extra data projection to improve the raw data distribution under complex view structures with more than two layers.
%In this paper, 
We have agglomerated converted subviews using only the weighted relationship, which has proven to achieve excellent performance. 
% \textcolor{red}{need to be more specific, what do you mean more complex? layers than two? and anything else?}
In light of the flexibility of ANN, we plan to extend ANN with more agglomerative relationships, e.g., convolutional networks, to solve general multi-view problems in more scenarios in the future. Besides, we will extend the proposed methods with matrix factorization to accelerate multi-view clustering.
%and agglomerative consensus analysis.
% \begin{algorithm}[tb]
%   \caption{Bubble Sort}
%   \label{alg:example}
% \begin{algorithmic}
%   \STATE {\bfseries Input:} data $x_i$, size $m$
%   \REPEAT
%   \STATE Initialize $noChange = true$.
%   \FOR{$i=1$ {\bfseries to} $m-1$}
%   \IF{$x_i > x_{i+1}$}
%   \STATE Swap $x_i$ and $x_{i+1}$
%   \STATE $noChange = false$
%   \ENDIF
%   \ENDFOR
%   \UNTIL{$noChange$ is $true$}
% \end{algorithmic}
% \end{algorithm}

\bibliographystyle{IEEEtran}

\bibliography{bio}

\begin{thebibliography}{10}
\providecommand{\url}[1]{#1}
\csname url@samestyle\endcsname
\providecommand{\newblock}{\relax}
\providecommand{\bibinfo}[2]{#2}
\providecommand{\BIBentrySTDinterwordspacing}{\spaceskip=0pt\relax}
\providecommand{\BIBentryALTinterwordstretchfactor}{4}
\providecommand{\BIBentryALTinterwordspacing}{\spaceskip=\fontdimen2\font plus
\BIBentryALTinterwordstretchfactor\fontdimen3\font minus
  \fontdimen4\font\relax}
\providecommand{\BIBforeignlanguage}[2]{{%
\expandafter\ifx\csname l@#1\endcsname\relax
\typeout{** WARNING: IEEEtran.bst: No hyphenation pattern has been}%
\typeout{** loaded for the language `#1'. Using the pattern for}%
\typeout{** the default language instead.}%
\else
\language=\csname l@#1\endcsname
\fi
#2}}
\providecommand{\BIBdecl}{\relax}
\BIBdecl

\bibitem{nie2016constrained}
F.~Nie, X.~Wang, M.~I. Jordan, and H.~Huang, ``The constrained laplacian rank
  algorithm for graph-based clustering,'' in \emph{Thirtieth AAAI Conference on
  Artificial Intelligence}, 2016.

\bibitem{vidal2011subspace}
R.~Vidal, ``Subspace clustering,'' \emph{IEEE Signal Processing Magazine},
  vol.~28, no.~2, pp. 52--68, 2011.

\bibitem{kriegel2011density}
H.-P. Kriegel, P.~Kr{\"o}ger, J.~Sander, and A.~Zimek, ``Density-based
  clustering,'' \emph{Wiley Interdisciplinary Reviews: Data Mining and
  Knowledge Discovery}, vol.~1, no.~3, pp. 231--240, 2011.

\bibitem{kleindessner2019guarantees}
M.~Kleindessner, S.~Samadi, P.~Awasthi, and J.~Morgenstern, ``Guarantees for
  spectral clustering with fairness constraints,'' \emph{arXiv preprint
  arXiv:1901.08668}, 2019.

\bibitem{oyelade2010application}
O.~Oyelade, O.~Oladipupo, and I.~Obagbuwa, ``Application of k means clustering
  algorithm for prediction of students academic performance,'' \emph{arXiv
  preprint arXiv:1002.2425}, 2010.

\bibitem{peng2020deep}
X.~Peng, J.~Feng, J.~T. Zhou, Y.~Lei, and S.~Yan, ``Deep subspace clustering,''
  \emph{IEEE Transactions on Neural Networks and Learning Systems}, 2020.

\bibitem{ergul2016clustering}
E.~Ergul, N.~Arica, N.~Ahuja, and S.~Erturk, ``Clustering through hybrid
  network architecture with support vectors,'' \emph{IEEE transactions on
  neural networks and learning systems}, vol.~28, no.~6, pp. 1373--1385, 2016.

\bibitem{aaai2017}
F.~Nie, G.~Cai, and X.~Li, ``Multi-view clustering and semi-supervised
  classification with adaptive neighbours,'' in \emph{Thirty-First AAAI
  Conference on Artificial Intelligence}, 2017.

\bibitem{abnomral2005}
Z.~Fu, W.~Hu, and T.~Tan, ``Similarity based vehicle trajectory clustering and
  anomaly detection,'' in \emph{IEEE International Conference on Image
  Processing 2005}, vol.~2.\hskip 1em plus 0.5em minus 0.4em\relax IEEE, 2005,
  pp. II--602.

\bibitem{ijcaisw}
Y.~Li, F.~Nie, H.~Huang, and J.~Huang, ``Large-scale multi-view spectral
  clustering via bipartite graph,'' in \emph{Twenty-Ninth AAAI Conference on
  Artificial Intelligence}, 2015.

\bibitem{peng2019comic}
X.~Peng, Z.~Huang, J.~Lv, H.~Zhu, and J.~T. Zhou, ``Comic: Multi-view
  clustering without parameter selection,'' in \emph{International Conference
  on Machine Learning}, 2019, pp. 5092--5101.

\bibitem{coreg2011}
A.~Kumar, P.~Rai, and H.~Daume, ``Co-regularized multi-view spectral
  clustering,'' in \emph{Advances in neural information processing systems},
  2011, pp. 1413--1421.

\bibitem{cotraining2011}
A.~Kumar and H.~Daum{\'e}, ``A co-training approach for multi-view spectral
  clustering,'' in \emph{Proceedings of the 28th International Conference on
  Machine Learning (ICML-11)}, 2011, pp. 393--400.

\bibitem{cikm17_mcge}
G.~Ma, L.~He, C.-T. Lu, W.~Shao, P.~S. Yu, A.~D. Leow, and A.~B. Ragin,
  ``Multi-view clustering with graph embedding for connectome analysis,'' in
  \emph{Proceedings of the 2017 ACM on Conference on Information and Knowledge
  Management}.\hskip 1em plus 0.5em minus 0.4em\relax ACM, 2017, pp. 127--136.

\bibitem{zhao2018incomplete}
L.~Zhao, Z.~Chen, Y.~Yang, Z.~J. Wang, and V.~C. Leung, ``Incomplete multi-view
  clustering via deep semantic mapping,'' \emph{Neurocomputing}, vol. 275, pp.
  1053--1062, 2018.

\bibitem{mvgl}
Z.~Kun, Z.~Changqing, G.~Junpeng, and W.~Junsheng, ``Graph learning for
  multiview clustering.'' \emph{IEEE transactions on cybernetics}, vol.~48,
  no.~10, p. 2887, 2018.

\bibitem{mvkdr}
X.~He, L.~Li, D.~Roqueiro, and K.~Borgwardt, ``Multi-view spectral clustering
  on conflicting views,'' in \emph{Joint European Conference on Machine
  Learning and Knowledge Discovery in Databases}.\hskip 1em plus 0.5em minus
  0.4em\relax Springer, 2017, pp. 826--842.

\bibitem{scikit-learn}
F.~Pedregosa, G.~Varoquaux, A.~Gramfort, V.~Michel, B.~Thirion, O.~Grisel,
  M.~Blondel, P.~Prettenhofer, R.~Weiss, V.~Dubourg, J.~Vanderplas, A.~Passos,
  D.~Cournapeau, M.~Brucher, M.~Perrot, and E.~Duchesnay, ``Scikit-learn:
  Machine learning in {P}ython,'' \emph{Journal of Machine Learning Research},
  vol.~12, pp. 2825--2830, 2011.

\bibitem{bmvc}
Z.~Zhang, L.~Liu, F.~Shen, H.~T. Shen, and L.~Shao, ``Binary multi-view
  clustering,'' \emph{IEEE transactions on pattern analysis and machine
  intelligence}, vol.~41, no.~7, pp. 1774--1782, 2018.

\bibitem{swmc}
F.~Nie, J.~Li, X.~Li \emph{et~al.}, ``Self-weighted multiview clustering with
  multiple graphs.'' in \emph{IJCAI}, 2017, pp. 2564--2570.

\bibitem{wang2019gmc}
H.~Wang, Y.~Yang, and B.~Liu, ``Gmc: Graph-based multi-view clustering,''
  \emph{IEEE Transactions on Knowledge and Data Engineering}, 2019.

\bibitem{zhang2015low}
C.~Zhang, H.~Fu, S.~Liu, G.~Liu, and X.~Cao, ``Low-rank tensor constrained
  multiview subspace clustering,'' in \emph{Proceedings of the IEEE
  international conference on computer vision}, 2015, pp. 1582--1590.

\bibitem{zhou2019multiple}
S.~Zhou, X.~Liu, M.~Li, E.~Zhu, L.~Liu, C.~Zhang, and J.~Yin, ``Multiple kernel
  clustering with neighbor-kernel subspace segmentation,'' \emph{IEEE
  transactions on neural networks and learning systems}, 2019.

\bibitem{andrew2013deep}
G.~Andrew, R.~Arora, J.~Bilmes, and K.~Livescu, ``Deep canonical correlation
  analysis,'' in \emph{International conference on machine learning}, 2013, pp.
  1247--1255.

\bibitem{wang2015deep}
W.~Wang, R.~Arora, K.~Livescu, and J.~Bilmes, ``On deep multi-view
  representation learning,'' in \emph{International Conference on Machine
  Learning}, 2015, pp. 1083--1092.

\bibitem{wei2019multi}
S.~Wei, J.~Wang, G.~Yu, X.~Zhang \emph{et~al.}, ``Multi-view multiple
  clusterings using deep matrix factorization,'' \emph{arXiv preprint
  arXiv:1911.11396}, 2019.

\bibitem{wu2019essential}
J.~Wu, Z.~Lin, and H.~Zha, ``Essential tensor learning for multi-view spectral
  clustering,'' \emph{IEEE Transactions on Image Processing}, vol.~28, no.~12,
  pp. 5910--5922, 2019.

\bibitem{zhang2018generalized}
C.~Zhang, H.~Fu, Q.~Hu, X.~Cao, Y.~Xie, D.~Tao, and D.~Xu, ``Generalized latent
  multi-view subspace clustering,'' \emph{IEEE transactions on pattern analysis
  and machine intelligence}, vol.~42, no.~1, pp. 86--99, 2018.

\bibitem{huang2019multi}
Z.~Huang, J.~T. Zhou, X.~Peng, C.~Zhang, H.~Zhu, and J.~Lv, ``Multi-view
  spectral clustering network,'' in \emph{Proceedings of the Twenty-Eighth
  International Joint Conference on Artificial Intelligence}.\hskip 1em plus
  0.5em minus 0.4em\relax International Joint Conferences on Artificial
  Intelligence Organization, 2019, pp. 2563--2569.

\bibitem{von2007tutorial}
U.~Von~Luxburg, ``A tutorial on spectral clustering,'' \emph{Statistics and
  computing}, vol.~17, no.~4, pp. 395--416, 2007.

\bibitem{lutkepohl1996handbook}
H.~L{\"u}tkepohl, \emph{Handbook of matrices}.\hskip 1em plus 0.5em minus
  0.4em\relax Wiley Chichester, 1996, vol.~1.

\bibitem{baruah2015dataset}
U.~Baruah and S.~M. Hazarika, ``A dataset of online handwritten assamese
  characters.'' \emph{Journal of Information Processing Systems}, vol.~11,
  no.~3, 2015.

\bibitem{lecun1998gradient}
Y.~LeCun, L.~Bottou, Y.~Bengio, and P.~Haffner, ``Gradient-based learning
  applied to document recognition,'' \emph{Proceedings of the IEEE}, vol.~86,
  no.~11, pp. 2278--2324, 1998.

\bibitem{demiriz2002linear}
A.~Demiriz, K.~P. Bennett, and J.~Shawe-Taylor, ``Linear programming boosting
  via column generation,'' \emph{Machine Learning}, vol.~46, no. 1-3, pp.
  225--254, 2002.

\bibitem{geusebroek2005amsterdam}
J.-M. Geusebroek, G.~J. Burghouts, and A.~W. Smeulders, ``The amsterdam library
  of object images,'' \emph{International Journal of Computer Vision}, vol.~61,
  no.~1, pp. 103--112, 2005.

\bibitem{fei2004learning}
L.~Fei-Fei, R.~Fergus, and P.~Perona, ``Learning generative visual models from
  few training examples: An incremental bayesian approach tested on 101 object
  categories,'' in \emph{2004 conference on computer vision and pattern
  recognition workshop}.\hskip 1em plus 0.5em minus 0.4em\relax IEEE, 2004, pp.
  178--178.

\bibitem{fanello2014dictionary}
S.~R. Fanello, N.~Noceti, G.~Metta, and F.~Odone, ``Dictionary based pooling
  for object categorization,'' in \emph{2014 International Conference on
  Computer Vision Theory and Applications (VISAPP)}, vol.~2.\hskip 1em plus
  0.5em minus 0.4em\relax IEEE, 2014, pp. 269--274.

\bibitem{hartmann2012consumer}
P.~Hartmann and V.~Apaolaza-Ib{\'a}{\~n}ez, ``Consumer attitude and purchase
  intention toward green energy brands: The roles of psychological benefits and
  environmental concern,'' \emph{Journal of business Research}, vol.~65, no.~9,
  pp. 1254--1263, 2012.

\bibitem{mandrik2005exploring}
C.~A. Mandrik and Y.~Bao, ``Exploring the concept and measurement of general
  risk aversion,'' \emph{ACR North American Advances}, 2005.

\end{thebibliography}




%\bibliographystyle{IEEEtran}

\end{document}